\documentclass[smallcondensed]{svjour3}       
\newcommand{\TheName}{\emph{De-Sparse}}
\usepackage[misc]{ifsym}

\usepackage{amssymb}
\usepackage{amsmath}
      \newtheorem{assumption}{Assumption}
\usepackage{graphicx} % for EPS, load graphicx instead
\usepackage{url}      % llt: nicely formatted URLs
\usepackage[lofdepth,lotdepth]{subfig}
\usepackage{multirow}
\usepackage{bbm}
\usepackage{pseudocode}
\usepackage{color}
\usepackage{multirow}
\usepackage{bbm}
\usepackage{pseudocode}
\usepackage{color}
\usepackage{algpseudocode}
\usepackage{booktabs}

\begin{document}

\authorrunning{Sijia Yang\and Haoyi Xiong\and Kaibo Xu\and Licheng Wang\and Jiang Bian\and Zeyi Sun}

\titlerunning{Improving Covariance-Regularized Discriminant Analysis for EHR-based Predictive.}

\title{Improving Covariance-Regularized Discriminant Analysis for EHR-based Predictive Analytics of Diseases}

\author{Sijia Yang\and Haoyi Xiong\and Kaibo Xu\and Licheng Wang\and Jiang Bian\and Zeyi Sun}

\date{Received: date / Accepted: date}

\institute{S. Yang \at
            School of Cyberspace Security, State Key Laboratory of Networking and Switching\\
            Beijing University of Posts and Telecommunications, Haidian, Beijing, China\\
            \email{ysjhhh@gmail.com}           %  \\
\and
           H. Xiong %(\Letter)
           \at
           Department of Computer Science\\
           Missouri University of Science and Technology, Rolla, MO\\
           \email{haoyi.xiong.fr@ieee.org}
\and
           K. Xu\at
            Mininglamp Academy of Sciences \\ 
            Mininglamp Technology, Beijing, 100084, China\\
            \email{xukaibo@mininglamp.com}
\and
           L. Wang~(\Letter)\at
            School of Cyberspace Security, State Key Laboratory of Networking and Switching\\
            Beijing University of Posts and Telecommunications, Haidian, Beijing, China\\
            \email{wanglc@bupt.edu.cn}
\and
            J. Bian\at
            Department of Electrical and Computer Engineering\\
            University of Central Floria, Orland, FL\\
            \email{bj1119@knights.ucf.edu}
\and
           Z. Sun (\Letter)\at
            Mininglamp Academy of Sciences \\
            Mininglamp Technology, Beijing, 100084, China\\
            \email{sunzeyi@mininglamp.com}
}
%\affil{Baidu Inc.}
%}
%\thanks{
%Author's addresses: 
%S. Yang and L. Wang are with State Key Laboratory of Networking and Switching Technology, Beijing University of Posts and Telecommunications, Beijing 100876, China; 
%Y. Zhang is with Department of  Computer Science, Missouri University of Science and Technology, Rolla MO 65401, United States;
%J. Bian and Z. Guo are with  Department of Electrical and Computer Engineering, University of Central Florida, Orlando FL 32816, United States;
%H. Xiong, Big Data Lab, Baidu Inc, Beijing 100193, China. 
%The first two authors contributed equally. 
%Please contact Prof. Licheng Wang and Dr. Haoyi Xiong for correspondence.
%}

\maketitle

%\IEEEcompsoctitleabstractindextext{

\begin{abstract}
Linear Discriminant Analysis (LDA) is a well-known technique for feature extraction and dimension reduction. The performance of classical LDA, however, significantly degrades on the High Dimension Low Sample Size (HDLSS) data for the \emph{ill-posed inverse problem}. Existing approaches for HDLSS data classification typically assume the data in question are with Gaussian distribution and deal the HDLSS classification problem with regularization. However, these assumptions are too strict to hold in many emerging real-life applications, such as enabling personalized predictive analysis using Electronic Health Records (EHRs) data collected from an extremely limited number of patients who have been diagnosed with or without the target disease for prediction. 

In this paper, we revised the problem of predictive analysis of disease using personal EHR data and LDA classifier. To fill the gap, in this paper, we first studied an analytical model that understands the accuracy of LDA for classifying data with arbitrary distribution. The model gives a theoretical upper bound of LDA error rate that is controlled by two factors: (1) the \emph{statistical convergence rate} of (inverse) covariance matrix estimators and (2) the divergence of the training/testing
datasets to fitted distributions. To this end, we could lower the error rate by balancing the two factors for better classification performance. Hereby, we further proposed a novel LDA classifier \TheName\ that leverages \emph{De-sparsified Graphical Lasso} to improve the estimation of LDA, which outperforms state-of-the-art LDA approaches developed for HDLSS data. Such advances and effectiveness are further demonstrated by both theoretical analysis and extensive experiments on EHR datasets.

%To understand the performance of Linear Discriminant Analysis (LDA) for HDLSS data classification,
%
%we studied a generalized framework for analyzing the misclassification rate of LDA on top of uncertain estimates of (inverse) covariance matrices under High Dimension Low Sample Size (HDLSS) settings.
%
%Specifically,
%we extend the existing LDA analytic model for multivariate Gaussian-distributed data classification,  and then obtained an upper bound of LDA misclassification rate for classifying data with arbitrary distribution (other than Gaussian), with respect to the uncertain estimates of (inverse) covariance matrices under HDLSS setting.
%

%To demonstrate the practical advances of proposed models, we evaluated the two LDA models with other regularized LDA and downstream classifiers, for early detection of diseases based on electronic health records (EHR) data. The evaluation result backups our theoretical analysis.
%
\end{abstract}

%}

%\IEEEpeerreviewmaketitle

\section{Introduction}
Linear Discriminant Analysis (LDA) \cite{Duda01a} is a well-known technique for feature extraction and dimension reduction. It has been widely used in many applications \cite{peck1982use,xiong2018mathcal} such as face recognition, image retrieval, etc. Typically, LDA finds the projection directions such that for the projected data, the between-class variance has been maximized relative to the within-class variance, thus achieving maximum discrimination. An intrinsic limitation of classical LDA is that its objective function requires the nonsingularity of one of the scatter matrices. For many applications, such as the microarray data analysis, all scatter matrices can be singular or ill-posed since the data is often with high dimension but low sample size (HDLSS)~\cite{buhlmann2011statistics}.

Recently, many efforts have been devoted to bear on such HDLSS problems. For example, Krzanowski et al. proposed a pseudo-inverse LDA to approximate the inverse covariance matrix, when the sample covariance matrix is singular. However, the accuracy of pseudo-inverse  LDA is usually low and not well guaranteed~\cite{krzanowski1995discriminant}. Another technique to alleviate this problem is a two-stage algorithm, \emph{i.e.}, PCA+LDA \cite{conf/eccv/BelhumeurHK96,Ye2004}. More popularly, regularized LDA approaches are proposed to solve the problem and improve the performance~\cite{tikhonov1943stability}. For example, researchers proposed a series of algorithms to regularize the covariance matrix estimation~\cite{peck1982use,krzanowski1995discriminant,witten2009covariance}. The regularized linear discriminant hyperplane was studied in~\cite{clemmensen2011sparse,shao2011sparse,buhlmann2011statistics}. All regularized LDA approaches intend to improve LDA through regularizing the estimation of key parameters used in LDA, such as the covariance matrix and/or the linear coefficients for discrimination.

One representative regularized LDA approach is {Covariance Regularized Discriminant Analysis} ({\em CRDA}) proposed in~\cite{witten2009covariance} based on the sparse inverse covariance estimation leveraging Graphical Lasso~\cite{friedman2008sparse}.
CRDA was originally proposed to estimate the inverse covariance matrix via a shrunken estimator, so as to achieve \emph{``superior prediction''}. Intuitively, through replacing the sample covariance matrix used in LDA with the regularized estimation, the HDLSS problem can be well addressed since the regularized estimators usually outperform the sample covariance matrix estimator~\cite{cai2016estimating}.
To better elucidate the performance of LDA classifiers with uncertain covariance matrix estimates for Gaussian data classification, ~\cite{zollanvari2013random} studied a model of error rate by matching the estimated \emph{vs}. true covariance matrices, and the estimated \emph{vs.} true means. While it is reasonable to assume that the estimated mean can easily converge to the population/true mean, the population/true covariance matrix is usually unknown and can be very different with the estimated one~\cite{cai2016estimating}. For example, the largest eigenvalue of the sample covariance matrix, which represents the principle component of the data distribution, is not consistent with the population one and the eigenvectors of the sample covariance matrix can be almost orthogonal to the truth under HDLSS~\cite{marvcenko1967distribution,johnstone2001distribution}.
Further, the data for classification is usually Non-Gaussian. Thus, it is highly desirable to
develop a new analytical model to characterize the error rate for the data with arbitrary distribution (both Gaussian and Non-Gaussian).
Two  ``known factors'' of covariance matrix estimation are useful for developing such analytical models, one is the convergence rate and the other one is the sparsity/density of (inverse) covariance matrix estimators~\cite{cai2016estimating}. The sparsity/density is already known once the (inverse) covariance matrix is estimated. The convergence rates reflect the maximal error of estimation, and for some estimators, they are well bounded under certain assumptions, such as spectral-norm convergence rate of  Graphical Lasso~\cite{rothman2008sparse}.

\begin{figure*}
\centering
\includegraphics[width=0.8\textwidth]{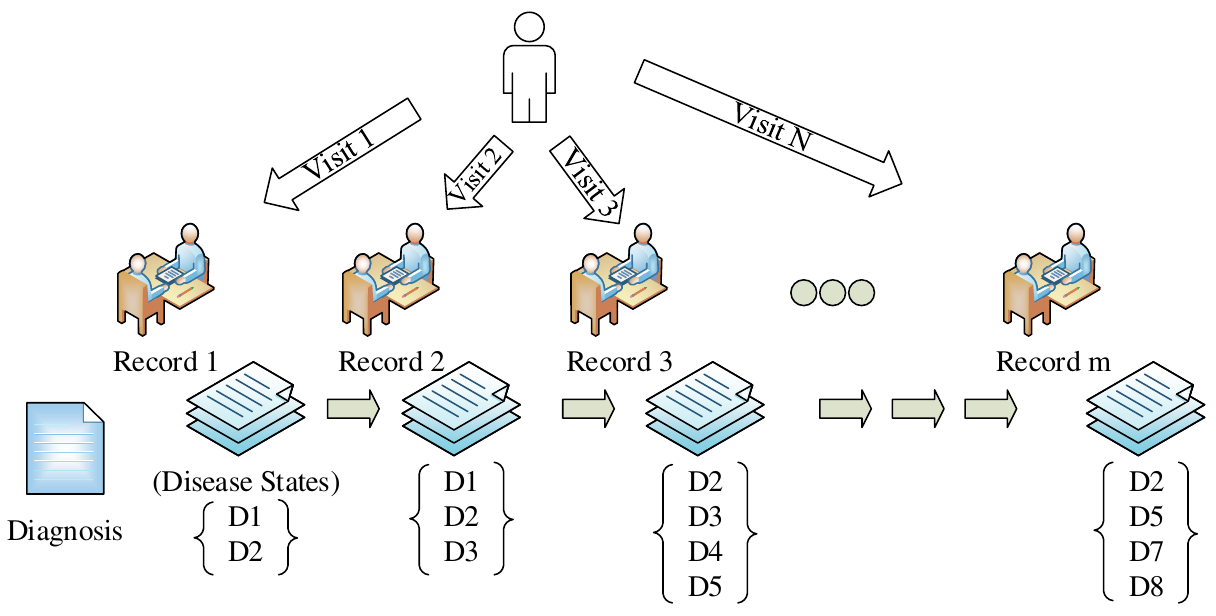}
\caption{Medicare visits and Electronic Health Records (EHRs). EHRs of a patient consist of the records of diagnoses and treatments. In this example, totally $m$ medical visits have been placed. For every medical visit, the patient would receive a set of ICD/CPT codes~\cite{yadav2018mining} referring the diseases/treatments that have been diagnosed/carried out. One can enable the early diagnosis/detection of diseases through classifying the EHR data, with big data and machine learning techniques.}
\label{fig:exp-ehr}
\end{figure*}

Among a wide range of HDLSS data classification tasks, in this work, we focus on the problem of using LDA to classify EHR~\cite{7091853} for personalized predictive analytics of target disease. EHRs play a critical role in modern health information management and service innovations.  {A patient's EHR contains his/her histories of medical visit, medication, diagnoses, treatment plans, allergies and so on as shown in Fig 1. Per each visit a diagnosis record would be updated indicating the disease state, i.e., a set of codes referring to the diseases that diagnosed at a time of visit.}
One significant feature is the interchangeability of EHR, as a standard protocol for medical/health data generation, storage and communication. The health information is built and managed by authorized institutions in a unified digital format (e.g., ICD-9/10, CPT-9/10 used in EHR standards) such that researchers and scientists can share and analyze the EHR data to enable innovative health services, such as providing computer-assisted diagnosis and offering medication advice. 
Among these services, predictive analytics of diseases (or namely early detection of diseases) using patients' past longitudinal health information of the EHR system, has recently attracted significant attention from research communities.
There has been a series of works \cite{sun2012supervised,7091853,personalized2015,zhang_mseq_2015,jensen2001mining,liu_temporal_2015,yang2018early,xiong2017daehr}, which attempt to predict future disease of patients, through data mining techniques using EHR data. Prior literature usually first selected important features, such as diagnosis-frequencies~\cite{7091853}, pairwise diagnosis transitions~\cite{zhang_mseq_2015}, and graphs of diagnosis sequences~\cite{liu_temporal_2015}, to represent the EHR data of the patients. Then, a wide range of supervised learning algorithms were adopted to build predictive models for early disease detection, on top of well-represented EHR data.

In this paper, we first propose a novel analytical model for LDA error rate, based on the statistical convergence of (inverse) covariance matrix estimators and the divergence to the Gaussian distributions. Guided by the proposed analytical model, we propose a novel LDA classifier leveraging the (inverse) covariance matrix estimators with faster convergence rate. We apply our model to a large-scale EHR dataset for the predictive analytics of diseases and demonstrate the advantage of the proposed algorithms over other state-of-the-arts. Specifically, in this paper, we made contributions as follows.
\begin{enumerate}

\item We studied the problem of high-dimensional linear classification using LDA models and proposed a novel analytical model, derived from the existing LDA models on Gaussian data~\cite{lachenbruch1968estimation,zollanvari2013random}, to understand the LDA error rate for both Gaussian and Non-Gaussian data, with respect to the statistical error of covariance matrices estimation and the divergence between fitted Gaussian distribution and the data. %To the best of our knowledge, this is the first study that understand the affects of statistical error for linear covariance models to the accuracy of classification, under the settings of EHR-based predictive analytics.

\item On top of the analytical model, we  proposed \TheName{}, which extends the well-known baseline approach -- \emph{Covariance Regularized Discrimiant Analysis (CRDA)}~\cite{witten2009covariance,bian2017early}, using De-sparsified Graphical Lasso~\cite{jankova2015confidence}. Theoretical analysis based on the proposed analytical model shows that \TheName\ can bound the maximal error rate, under mild conditions. Compared to CRDA, \TheName\ could achieve lower error rate, due to the faster convergence rate of De-sparsified Graphical Lasso.

\item To show the practical contribution of the proposed method, we evaluated \TheName{} extensively through experiments with large-scale, real-world EHR datasets~\cite{turner_college_2015}. In the experiments, we used \TheName\ to predict the risk of mental health disorders in college students from ten U.S. universities, using their EHR data from primary care visits. We compared \TheName\ with seven baseline algorithms including other regularized LDA models and downstream classifiers. The evaluation result shows that \TheName\ outperforms all baselines, and  further validates our theoretical analysis.
\end{enumerate}
The paper is organized as follows. In Section 2, we review the problem of high-dimensional linear classification using LDA models and summarize the existing work on EHR-based predictive analytics of diseases. In Section 3, we first introduce the existing covariance-regularized discriminant analysis (CRDA) based on Graphical Lasso, then present de-sparsified covariance regularized LDA algorithms, based on novel de-sparsified inverse covariance matrix estimators, to classify EHR samples for the predictive analytics. In Section 4, we validate the proposed algorithms with real-world datasets through extensive experiments. Finally, we conclude the paper in Section 5.

%For the rest of this paper, we first review LDA classifiers for binary classification, along with the state of the art of the LDA error rate analysis (for Gaussian data only), in Section 2. Then, we present the main theory of this paper -- a new analytic model characterizing LDA error upper bounds (for both Gaussian and Non-Gaussian data), and the proposed algorithm -- a novel LDA classifier  \TheName\ with error upper bound discussed, in Section 3 and 4 respectively. Finally, we present the experiment results comparing \TheName\ with baseline algorithms, then conclude the paper, in Section 5 and 6 respectively.

%The rest of this paper is arranged as follow: Section 2 introduces the preliminaries of this study and formulate the research problem; Section 3 presents the proposed algorithm and Section 4 analyzes the performance of proposed solution theoretically. We addressed the setups of our experiment as well as the results in Section 5. Section 6 and 7 discuss the most relevant work and conclude this paper, respectively.

%present \TheName{}---\emph{Linear Discriminant Analysis (LDA) Model with Few Patients' EHR Data}---an extension to LDA~\cite{fisher1936use,mclachlan2004discriminant} framework for early detection of diseases using Electronic Health Records (EHR), which can improve the prediction accuracy of the standard LDA model by reducing the noise in EHR data and regularizing the estimated covariance matrices.
%Specifically, \TheName\  Finally, we conducted . We compared our solution with other regularized LDA and downstream classifiers.

\section{Preliminaries}
%In this section, we first brief the binary classifier using traditional linear discriminant analysis (LDA). Then, we present the state-of-the-art of analytical models on the LDA error rate that assume the data for classification follow certain Gaussian distributions.

\subsection{LDA for Binary Classification}
To use Fisher's Linear Discriminant Analysis (FDA), given $N$ labeled data pairs $(x_1,l_1), (x_2,l_2),(x_3,l_3)\dots$ $(x_N,l_N)$ and $\forall x_i$, $1\leq i\leq N$ refers to a $d$-dimensional vector, we first estimate the sample covariance matrix (an symmetric $d\times d$ matrix) using maximized likelihood estimator:
\begin{equation}
\begin{aligned}
\bar\Sigma&=\frac{1}{N}\sum_{i=1}^{N}( x_i-\bar{\mu})(x_i-\bar{\mu})^\top,
\end{aligned}
\label{eq:sample-cov}
\end{equation}
where $\bar{\mu}$ refers to the $d$-dimensional mean vector of all $N$ training samples $(x_1,l_1),$ $(x_2,l_2)\dots(x_N,l_N)$. Then, $\bar\mu_+$ and $\bar\mu_-$ are estimated as the mean vectors of the positive training samples and negative training samples in the $N$ training samples, respectively.

\begin{corollary}[Fisher's Discriminant Analysis for Binary Classification~\cite{Duda01a}]
Given all estimated parameters $\bar\Sigma$, $\bar\mu_+$, and $\bar\mu_-$, the FDA model classifies a new data vector $x$ as the result of Eq.~\ref{eq:gLDA} as follows.
\begin{equation}
\begin{aligned}
f_{\bar\Sigma}(x)=sign \left(log \frac{x^\top\bar\Sigma^{-1}\bar\mu_+-\frac{1}{2}\bar\mu_+^\top\bar\Sigma^{-1}\bar\mu_++\log\pi_+}{x^\top\bar\Sigma^{-1}\bar\mu_--\frac{1}{2}\bar\mu_-^\top\bar\Sigma^{-1}\bar\mu_-+\log\pi_-} \right),
\end{aligned}
\label{eq:gLDA}
\end{equation}
where $sign(\cdot):\mathbbm{R}\to \{\pm 1\}$ refers to the signal function, $\pi_+$ and $\pi_-$ refer to the (foreknown) frequencies of positive samples and negative samples in the whole population.
\end{corollary}
To present LDA with other covariance matrix estimator, based on the LDA paradigm listed in Eq.~\ref{eq:gLDA}, we use the notations as follows.

\noindent\textbf{Notations}. \emph{Note that, in the rest of this paper, we denote $f_{\widehat \Sigma}(x)$ as an LDA classifier with a specific covariance matrix estimator $\widehat\Sigma$, using the sample estimated mean vectors $\bar\mu_-$ and $\bar\mu_+$.
%\begin{equation}
%\begin{aligned}
%f_{\widehat\Sigma}=sign \left(log \frac{x^\top\widehat\Sigma^{-1}\bar\mu_+-\frac{1}{2}\bar\mu_+^\top\widehat\Sigma^{-1}\bar\mu_++\log\pi_+}{x^\top\widehat\Sigma^{-1}\bar\mu_--\frac{1}{2}\bar\mu_-^\top\widehat\Sigma^{-1}\bar\mu_-+\log\pi_-} \right).
%\end{aligned}
%\label{eq:cLDA}
%\end{equation}
%
When $\widehat\Sigma=\bar\Sigma$, then the classifier $f_{\bar\Sigma}(x)$  becomes the traditional Fisher's Linear Discriminant Analysis. When $\widehat\Sigma=\widehat\Theta^{-1}$ and $\widehat\Theta$ is the Graphical Lasso estimator ~\cite{friedman2008sparse}, then  $f_{\widehat\Theta^{-1}}(x)$ refers to the \emph{covariance regularized LDA}~\cite{witten2009covariance,bian2017early}.}
%\end{notation}

Apparently, the performance of LDA depends on (1) whether the realistic training/testing datasets follow Gaussian distributions and (2) how the mean vectors and inverse covariance matrices are estimated from the datasets.

\subsection{Electronic Health Records and Predictive Analytic of Disease}

%\begin{figure}
%    \centering
%    \includegraphics[width=0.85\textwidth]{Patient.png}
%    \caption{Electronic Health Record (EHR) raw data}
%    \label{fig:patients}
%\end{figure}

%

Prior to learning a predictive analytic model for certain diseases, one needs to model the EHR data with a suitable data representation. The most simple yet effective way to represent EHR data is to use \emph{diagnosis-frequency vector}~\cite{van2010corpus,yu2011performance,shen2017populating}, which is similar to Term Frequency (TF) or Term Frequency-Inverse Document Frequency (TF-IDF) approach that deals with traditional NLP data~\cite{luhn1957statistical,jones1972statistical,aizawa2003information}. Given each patient's EHR data (shown in Fig~\ref{fig:exp-ehr}), this representation method first retrieves the diagnosis codes~\cite{dubberke2006icd} recorded during each visit. Inspired by some Natural Language Processing (NLP) and text mining practices~\cite{kowsari2019text}, researchers also proposed using some deep learning based NLP approaches to embed EHR records for data representation learning~\cite{choi2016multi,zhang2018patient2vec,choi2017gram,bai2018interpretable,ma2018health,rajkomar2018scalable}. For example,~\cite{zhang2018patient2vec} studied ``Patient2Vec'' which embeds patients' past EHR records into vectors while preserving structural information for personalized predictive analysis.~\cite{bai2018interpretable} focuses on the interpolation and interpretability of EHR representation learning, where authors well-balanced the performance of predictive analysis and the understanding to the longitude disease progress of each individual patient, both using the EHR data with the learned representation. Comprehensive surveys could be found in~\cite{shickel2017deep,yadav2018mining,solares2020deep}.

In our work, we follow the line of research that uses \emph{diagnosis-frequency vector} of each patient for EHR-based predictive analysis~\cite{van2010corpus,yu2011performance,shen2017populating}, as the diagnosis-frequency in a certain duration could well characterize the health status of patients while the coefficient of LDA can represent the significance of every diagnosis code. The frequency of each diagnosis appearing in all past visits (of the last two years) is counted here, followed by further transformation on the frequency of each diagnosis into a vector of frequencies.
For example, a \emph{diagnosis-frequency vector} can be illustrated as $\langle 1, 0, \dots, 3\rangle$, where 0 means the second diagnosis does not exist in all past visits. Note that the dimension $d\geq 15,000$ when using original ICD-9 codes, $d = 295$ even when using clustered ICD-9 codes~\cite{clusteredcodes}, while the number of samples for training $N$ in our experiment is significantly smaller than $d$.

\subsection{Discussion on Preliminaries}
In our work, we revisited the linear discriminant analysis as a classifier and learner for High-Dimensional and Low Sample Size (HDLSS) settings. Indeed, many efforts have been made in the literature for HDLSS data classification. For example, in addition to LDA-type methods, a number of feature extraction or variable selection methods have been studied~\cite{xiong2017awda,xiong2018mathcal}. Lin et al.,~\cite{sun2019feature} proposed a  feature selection algorithm to classify the high-dimensional gene expression data through incorporating the neighborhood entropy-based uncertainty measures. Over the rough set, the same group of authors~\cite{sun2019joint} adopted a joint feature selection approach that incorporates neighborhood entropy and the fisher scores, for tumor classification. Further, some automatic feature weighting paradigm has been proposed to select features for gene expression data classification~\cite{chen2012automated}. These studies demonstrate that the feature selection algorithms could significantly improve the accuracy of HDLSS data classification, while avoiding the full set of features. The over-reduction problem of LDA has been studied in~\cite{wan2017separability}. In addition to the EHR data, similar regularized projection methods have been used for early diagnosis of diseases for biomedical health data~\cite{yang2020inverse,xiao2019multi}.

In terms of methodologies, the most close work to this study is covariance-regularized linear discriminant analysis (CRDA)~\cite{witten2009covariance}, Graphical Lasso~\cite{rothman2008sparse}, and the de-sparsified Graphical Lasso~\cite{jankova2015confidence,xiong2018biasing}. CRDA regularizes the estimation of (inverse) covariance matrices inside the estimation of LDA, while improving the performance of LDA for both prediction and inference. Authors in~\cite{bian2017early} were the first to bring CRDA for EHR classification and early detection of diseases. We included the algorithms in~\cite{bian2017early} for comparison and found that \TheName\ outperformed CRDA with higher accuracy and F1-score. Compared to the Graphical Lasso estimator~\cite{rothman2008sparse} that has been frequently used to enhance the inverse covariance matrices estimation, our work followed the ideas of de-biased estimator~\cite{marozzi2015multivariate} and used de-sparsified Graphical Lasso estimator~\cite{jankova2015confidence} to improve the LDA for EHR classification. We would provide a comprehensive discussion on the performance comparisons between Graphical Lasso and de-sparsified one from the perspectives of predictive analytics based on EHR data and LDA.

\section{\TheName:  De-Sparsified Covariance-Regularized Discriminant Analysis}%: Model and Analysis}
%\subsection{Inverse Covariance Matrix Estimation Using De-Sparsified Graphical Lasso}
In this section, we first introduce the baseline algorithm based on Covariance-Regularized Discriminant Analysis (CRDA) that is derived from~\cite{bian2017early}. Then, we present the proposed algorithm \TheName, an extended Covariance Regularized Discriminant Analysis via De-sparsified Graphical Lasso~\cite{jankova2015confidence}. Then, using our proposed analytical model of LDA error rate, we compare two methods and demonstrate the advantages of \TheName.

\subsection{CRDA: The Baseline of Covariance-Regularized Discriminant Analysis via Graphical Lasso Inverse Covariance Matrix Estimator}
Compared to the sample LDA introduced in Section 2, CRDA~\cite{witten2009covariance,bian2017early} was proposed to use $\ell_1$-penalized inverse covariance matrix estimator to replace the inverse of sample covariance matrix. Given the labeled data pairs for training $(x_1,l_1)$, $(x_2,l_2)\dots$ $(x_{N},l_{N})$, the algorithm first estimates the sample covariance matrix $\bar\Sigma$ and the sample mean vectors $\bar\mu_+$, $\bar\mu_-$ using the algorithms introduced in Section 2.1.
With the sample covariance matrix $\bar\Sigma$, this method estimates a sparse inverse covariance matrix $\widehat\Theta$  using the Graphical Lasso estimator~\cite{friedman2008sparse} as follows.
\begin{corollary}[Graphical Lasso Estimator~\cite{friedman2008sparse}] Given the sample estimation of the covariance matrix $\bar\Sigma$, the Graphical Lasso estimator provides an $\ell_1$-regularized sparse positive-definite approximation to the inverse covariance matrix (denoted as $\widehat\Theta$) as follows.
\begin{equation}
{\displaystyle {\widehat {\Theta }}={\underset {\Theta>0}{\operatorname {argmin} }}\!\left(\operatorname{tr} (\bar\Sigma\Theta
)-\log \det(\Theta )+\lambda \sum _{j\neq k}{|\Theta _{jk}|}\right)},\label{eq:glasso}
\end{equation}
where $\Theta>0$ refers to the constraint of symmetric positive definiteness (SPD), the term $\operatorname{tr} (\bar\Sigma\Theta
)-\log \det(\Theta )$ refers to the negative log-likelihood of the optimization objective $\Theta$ over the sample estimate $\bar\Sigma$, the term $\sum _{j\neq k}{|\Theta _{jk}|}$ refers to the sum of absolute values of the non-diagonal elements in the matrix $\Theta$ (which is the same as the $\ell_1$-norm of $\Theta$ without diagonal elements considered), and $\lambda$ refers to tuning parameter that makes trade-off between the sparsity and the fitness to samples. Please refer to~\cite{friedman2008sparse} for the implementation of the algorithms.
\end{corollary}

\begin{corollary}[Statistical Convergence of Graphical Lasso~\cite{rothman2008sparse}] Suppose the random vector $\mathbf{X}$ is with $d$-dimensions and zero mean (i.e., $\mathbf X\in\mathbb{R}^d$ and $\mathbb E\ (\mathbf{X}) = \mathbf{0}$), where the population estimate of the covariance matrix is $\Sigma=\mathbb{E}\ (\mathbf{XX}^\top)$ and the inverse of population covariance matrix is $\Theta=\Sigma^{-1}$. Given $N$ samples $x_1,x_2,x_3,\dots,x_N$ randomly and independently drawn from $\mathbf{X}$, the sample estimate of the covariance matrix here should be $\bar\Sigma=\frac{1}{N}\sum_{i=1}^{N}x_ix_i^\top$. 

With the increasing number of samples ($N$) given and growing number of dimensions of the data ($d$), the graphical lasso estimate $\widehat\Theta$ based on the sample covariance matrix converges to the population estimate $\Theta$ at the rate of Frobenius-norm under mild sparsity conditions, as follows ~\cite{rothman2008sparse}.
%$\bar\Sigma$
\begin{equation}
    \|\widehat\Theta-\Theta\|_\mathrm{F}=\mathcal{O}\left(\sqrt{\frac{(d+s)\ \log p}{N}}\right),
\end{equation}
where $s=\mathrm{max}_{1\leq i\leq d}\|\Theta_i\|_0$ refers to the maximal degree of the graph in $\Theta$, $\|\cdot\|_0$ refers to the $\ell_0$-norm of the input vector, and $\Theta_i$ refers to the $i^{th}$ column  vector ($1\leq i\leq d$) of the matrix $\Theta_i$.
\end{corollary}

%
%Above estimator can be considered as a $\ell_1$-penalized negative log-likelihood minimization estimator, which provides an sparse approximation to the inverse of sample covariance matrix. In our work, we use Graphical Lasso~\cite{friedman2008sparse} to implement the estimator in Eq.~\ref{eq:glasso}. 
To the end, the classification rule of CRDA is written as follows
\begin{equation}
\begin{aligned}
\mathrm{CRDA}(x)=sign \left(log \frac{x^\top\widehat\Theta\bar\mu_+-\frac{1}{2}\bar\mu_+^\top\widehat\Theta\bar\mu_++\log\pi_+}{x^\top\widehat\Theta\bar\mu_--\frac{1}{2}\bar\mu_-^\top\widehat\Theta\bar\mu_-+\log\pi_-} \right),
\end{aligned}
\label{eq:crda}
\end{equation}
which can be viewed as an LDA classifier using $\widehat\Theta^{-1}$ as the covariance matrix. Apparently, the accuracy of CRDA depends on how the covariance matrices and the mean vectors are estimated. We are going to interpret the performance of CRDA in the Section 3.3. %the statistical convergence $\widehat\Theta$ to $\Sigma^{*-1}$.

\subsection{\TheName: The Improved Algorithm of Covariance-Regularized LDA via De-Sparsified Graphical Lasso}
As shown in Eq.~\ref{eq:glasso}, the estimator of sparse inverse covariance matrix induced $\ell_1$-penalization and might hurt the estimation due to the over-penalization or over-sparsification. To address this issue, we proposed a de-sparsified Graphical Lasso estimator~\cite{jankova2015confidence} to replace the vanilla Graphical Lasso.% in the classification rule, as follow. 

\begin{corollary}[De-sparsified Graohical Lasso~\cite{jankova2015confidence}] Given the Graphical Lasso estimator $\widehat\Theta$ and the sample estimation $\bar\Sigma$, we consider the inverse of Graphical Lasso $\widehat\Theta^{-1}$ as an approximation to the covariance matrix. In this way, the bias of $\widehat\Theta^{-1}$, caused by the sparsity regularizer of Graphical Lasso, for covariance estimation could be written as follows.
\begin{equation}
     \widehat Z=\bar\Sigma - \widehat\Theta^{-1}.
\end{equation}
%Suppose $\Sigma$ and $\Theta=\Sigma^{-1}$ are denoted as the population covariance matrix and its inverse. There exists
%\begin{equation}
%     \widehat\Theta-\Theta =  \Theta\left(\bar\Sigma-\Sigma\right)\Theta +\widehat\Theta\widehat Z\widehat\Theta
%\end{equation}
%
Using the Kronecker product, authors in~\cite{jankova2015confidence} consider the potential bias term of $\widehat\Theta$ against the inverse of population covariance matrix as follows.
\begin{equation}
    \mathrm{Bias}(\bar\Sigma,\widehat\Theta) = \widehat\Theta\widehat Z\widehat\Theta=\widehat\Theta\bar\Sigma\widehat\Theta - \widehat\Theta.
\end{equation}
The de-sparsified Graphical Lasso estimator $\widehat T$ de-biases the Graphical Lasso estimator $\widehat\Theta$ through removing the potential bias term caused by the sparsity reguarlizer, as follows.
\begin{equation}
\widehat T=\widehat\Theta- \mathrm{Bias}(\bar\Sigma,\widehat\Theta)=2\widehat\Theta-\widehat\Theta\bar\Sigma\widehat\Theta.\label{eq:de-sparse}
\end{equation}
On top of the Graphical Lasso, the de-sparsified Graphical Lasso estimator can efficiently approximate an estimation of inverse covariance matrix using the data with \emph{faster convergence rate} in a mild condition.
\end{corollary}

\begin{corollary}[Statistical Convergence of De-sparsified Graphical Lasso~\cite{jankova2015confidence}] Suppose the random vector $\mathbf{X}$ is with $d$-dimensions and zero mean (i.e., $\mathbf X\in\mathbb{R}^d$ and $\mathbb E\ (\mathbf{X}) = \mathbf{0}$), where the population estimate of the covariance matrix is $\Sigma=\mathbb{E}\ (\mathbf{XX}^\top)$ and the inverse of population covariance matrix is $\Theta=\Sigma^{-1}$. Given $N$ samples $x_1,x_2,x_3,\dots,x_{N}$ randomly and independently drawn from $\mathbf{X}$, the sample estimate of the covariance matrix here should be $\bar\Sigma=\frac{1}{m}\sum_{i=0}^{m-1}x_ix_i^\top$. The Graphical Lasso estimator and the de-sparsified estimator are denoted as $\widehat\Theta$ and $\widehat T$, respectively.

With the increasing number of samples ($N$) given and growing number of dimensions of the data ($d$), the De-sparsified Graphical Lasso estimator $\widehat T$ converges to the population estimate $\Theta$ at the rate of $\ell_\infty$-norm under mild sparsity conditions, as follows~\cite{jankova2015confidence}.
%$\bar\Sigma$
\begin{equation}
    \|\widehat\Theta-\Theta\|_\infty=\mathcal{O}\left(\sqrt{\frac{\log d}{N}}\right).
\end{equation}
Note that, above convergence rate of de-sparsified Graphical Lasso was obtained under similar sparsity assumptions as~\cite{rothman2008sparse}, while the $\ell_2$-norm or $\ell_\infty$-norm convergence rates of Graphical Lasso remain unknown yet.
\end{corollary}

Based on \emph{Notations}, we denote the De-sparsified Covariance Regularized Discriminant Analysis (namely \TheName) as $\mathrm{Desparse}(x)$, using the De-sparsified Graphical Lasso $\widehat T$ and the mean vectors $\bar\mu_+$, $\bar\mu_-$.
\begin{equation}
\begin{aligned}
\mathrm{Desparse}(x)=sign \left(log \frac{x^\top\widehat T\bar\mu_+-\frac{1}{2}\bar\mu_+^\top\widehat T\bar\mu_++\log\pi_+}{x^\top\widehat T\bar\mu_--\frac{1}{2}\bar\mu_-^\top\widehat T\bar\mu_-+\log\pi_-} \right).
\end{aligned}
\label{eq:dlda}
\end{equation}
With the de-sparsified inverse covariance matrix estimator $\widehat T$ enjoying better statistical properties, \TheName\ is expected to outperform CRDA with better classification accuracy. Detailed comparison will be discussed in the following sections.

\subsection{Performance Analysis of LDA, CRDA, and \TheName}
In this section, we first review the previous studies on the LDA error rate estimation for Gaussian data~\cite{lachenbruch1968estimation,zollanvari2013random}, then we generalize LDA error rate from Gaussian data to non-Gaussian data. Finally, we provide a discussion on the classification accuracy comparison among vanilla LDA, CRDA, and \TheName. 

\subsubsection{LDA Error Rate for Gaussian Data via Random Matrix Theory}
%In this section, we summarize the studiesin theoretical error rate of LDA classifiers for classifying multivariate Gaussian data.

We first assume the data for binary classification follow two (unknown) Gaussian distributions with the same covariance matrix $\Sigma$ but two different means $\mu_+$ and $\mu_-$, i.e., $\mathcal{N}(\mu_+,\Sigma)$ for positive samples and $\mathcal{N}(\mu_-,\Sigma)$ for negative samples, respectively.
Given the LDA classifier $f_{\widehat\Sigma}(x)$ based on the sample estimated mean vectors $\bar\mu_-$, $\bar\mu_+$ and a specific covariance matrix $\widehat\Sigma$, the expected error rate of a linear discriminant analysis (i.e., probability of  $l\neq f_{\widehat\Sigma}(x)$) on the  data of $\mathcal{N}(\mu_+,\Sigma)$, $\mathcal{N}(\mu_-,\Sigma)$   is modeled as follows.

\begin{corollary}[RMT-based LDA Error Rate Estimation~\cite{zollanvari2013random}] According to the random matrix theory, \cite{zollanvari2013random} models the expectation of classification error rate of LDA (using estimated parameters $\bar\mu_+$, $\bar\mu_-$, and $\widehat\Sigma$) on Gaussian distributions $\mathcal{N}(\mu_+,\Sigma)$ and $\mathcal{N}(\mu_-,\Sigma)$ as 
$\varepsilon(\bar\mu_+,\bar\mu_-,\widehat\Sigma,\mu_+,\mu_-,\Sigma)$, as follows.  
  \begin{equation}
	\begin{aligned}
\varepsilon(\bar\mu_+,\bar\mu_-,\widehat\Sigma,\mu_+,\mu_-,\Sigma) 
	&=\pi_+\cdot\Phi\left(-\frac{(\mu_+-\frac{(\bar\mu_++\bar\mu_-)}{2})^\top\widehat\Sigma^{-1} (\bar\mu_+-\bar\mu_-)}{\sqrt{(\bar\mu_+-\bar\mu_-)^\top\widehat\Sigma^{-1} \Sigma\widehat\Sigma^{-1} (\bar\mu_+-\bar\mu_-)}}\right)\\
	&+\pi_-\cdot\Phi\left(\frac{(\mu_--\frac{(\bar\mu_++\bar\mu_-)}{2})^\top\widehat\Sigma^{-1} (\bar\mu_+-\bar\mu_-)}{\sqrt{(\bar\mu_+-\bar\mu_-)^\top\widehat\Sigma^{-1} \Sigma\widehat\Sigma^{-1} (\bar\mu_+-\bar\mu_-)}}\right)
	\end{aligned}
		\label{eq:errorrate}
  \end{equation}
where $\Phi$ refers to the CDF function of a standard normal distribution.
\end{corollary}

According to \textbf{Corollary 6} and~\cite{zollanvari2013random}, we could conclude that the expected error rate is sensitive with the parameters $\mu_+,\mu_-,\Sigma,\widehat\mu_+,\widehat\mu_-$ and $\widehat\Sigma$, while the true parameters $\mu_+,\mu_-,\Sigma$ are assumed unknown. Compared to the (inverse) covariance matrices estimation, the error of sample mean vector estimation is relatively small~\cite{field1982small}. Thus, we adopt the settings in  studies~\cite{lachenbruch1968estimation,peck1982use,krzanowski1995discriminant} as follows.

\begin{assumption}
In this paper, we make \emph{\bf no assumptions} on the mean vectors $\mu_+,\mu_-,\mu$ and always use the sample mean  $\bar\mu_+,\bar\mu_-,\bar\mu$ to estimate $\mu_+,\mu_-,\mu$.
%\end{assumption}
%
%
%%\begin{assumption}
Even under the HDLSS settings, with  a certain number of samples,  it is reasonable to assume the sample estimation of mean vectors $\bar\mu_+$ and $\bar\mu_-$  should be close to the population mean vectors, i.e., $|\mu_+-\bar\mu_+|\to 0$, $|\mu_--\bar\mu_-|\to 0$, and $|\mu-\bar\mu|\to 0$.
\end{assumption}
\begin{lemma}
Thus, based on \textbf{Theorem 1} and the sample mean relaxation (\textbf{Assumption 1}), the expected error rate of $f_{\widehat\Sigma}(x)$ can be reduced to
\begin{equation}
	\begin{aligned}
	\varepsilon(\widehat\Sigma,\Sigma)
	=\Phi\left(-\frac{(\bar\mu_+-\bar\mu_-)^\top\widehat\Sigma^{-1} (\bar\mu_+-\bar\mu_-)}{2\sqrt{(\bar\mu_+-\bar\mu_-)^\top\widehat\Sigma^{-1} \Sigma\widehat\Sigma^{-1} (\bar\mu_+-\bar\mu_-)}}\right).
	\end{aligned}
		\label{eq:r-errorrate}
  \end{equation}
In this way, to improve the LDA classifier with the sample mean vectors, there needs an estimator $\widehat\Sigma$ to minimize or lower the expected error rate $\varepsilon(\widehat\Sigma,\Sigma)$.
\end{lemma}
\begin{lemma}
Furthermore, when the estimated covariance matrix $\widehat\Sigma$ is set to the oracle one $\Sigma$ (the LDA is perfectly fitted with the data), the expected error rate reaches the optimal error rate 
\begin{equation}
\varepsilon(\widehat\Sigma,\widehat\Sigma)=\Phi\left(-\frac{\sqrt{(\bar\mu_+-\bar\mu_-)^\top\widehat\Sigma^{-1} (\bar\mu_+-\bar\mu_-)}}{2}\right).   
\label{eq:optimal-rate}
\end{equation}
Above result suggests that when the covariance matrix $\widehat\Sigma$ is perfectly estimated $\widehat\Sigma\to\Sigma$ and $\widehat\Sigma^{-1}\Sigma\to I$, the LDA classifier would approach to its optimal error rate.
\end{lemma}
The estimate in Eq.~\ref{eq:r-errorrate} reduces the estimation of LDA classification error rate to the divergence between the population covariance matrix $\Sigma$ and the estimated one $\widehat\Sigma$. On the other hand, Eq.~\ref{eq:optimal-rate} models the generalization error of a model with ``perfectly-fitted'' covariances~\cite{zollanvari2013random}.

\subsubsection{Performance Comparisons}
%In this section, we intend to understand the performance of \TheName\  $f_{\widehat T^{-1}}(x)$, through comparing it with the existing CRDA  $f_{\widehat\Theta^{-1}}(x)$. According to our proposed analytical model, the performance of LDA highly depends on the \emph{convergence rate} of the (inverse) covariance matrix estimation and the divergence between fitted Gaussian models to  data.%, as well as its \emph{sparsity (or density)},  in $\ell_2$-norm.

%Hence, we find following \textbf{\em trade-off between convergence rate and sparsity}:
%\begin{itemize}

%\item
%\emph{Convergence Rate Comparison - } 
We compare the convergence rate of $\widehat T$ and $\widehat\Theta$ to the inverse population covariance matrix $\Theta^*=\Sigma^{*-1}$, so as to understand the accuracy of \TheName\ and CRDA. Since the divergence of the datasets to the nearest Gaussian distributions should be the same for both algorithms, we mainly compared the Gaussian error terms for \TheName\ and CRDA, i.e., $\varepsilon(\widehat T^{-1},\Sigma^*)$ vs. $\varepsilon(\widehat\Theta^{-1},\Sigma^*)$.  More specifically,~\cite{zollanvari2013random} demonstrated the connections between the Gaussian data error terms and the spectral properties of $\widehat T$ and $\widehat\Theta$. Considering \textbf{Lemma 2}, we hope to understand (1) how close the matrices $\widehat T\Sigma^*$ and $\widehat\Theta\Sigma^*$ would approach to $I$ matrix and (2) how the spectrum of these matrices behaves~\cite{zollanvari2011analytic}, such that
\begin{equation}
    \begin{aligned}
    \|\widehat T\Sigma^*-I\|_2& =\|(\widehat T-\Theta^*)\Sigma^*\|_2\leq\lambda_\mathrm{max}(\Sigma^*)\|\widehat T-\Theta^*\|_2,\ \text{and} \\
    \|\widehat\Theta\Sigma^*-I\|_2 & =\|(\widehat\Theta-\Theta^*)\Sigma^*\|_2\leq\lambda_\mathrm{max}(\Sigma^*)\|\widehat\Theta-\Theta^*\|_2,
    \end{aligned}
\end{equation}
where $\lambda_\mathrm{max}(\cdot)$ refers to the largest eigenvalue of the input matrix. Obviously, the terms of $\lambda_\mathrm{max}(\Sigma^*)$, $\|\widehat T-\Theta^*\|_2$ and $\|\widehat\Theta-\Theta^*\|_2$ are non-negative. When $\|\widehat T-\Theta^*\|_2\to 0$ and $\|\widehat\Theta-\Theta^*\|_2\to 0$, then optimal error rates would be achieved asymptotically. In this way, we are wondering whether $\widehat T$ would converge to $\Theta^*$ faster than $\widehat\Theta$ in the spectrum-norm distance.

Considering \textbf{Corollaries 3} and \textbf{5}, the sharp spectrum-norm statistical convergence rate of Graphical Lasso is not known in any of the previous studies~\cite{rothman2008sparse,banerjee2008model,jankova2015confidence,cai2016estimating}, while the spectrum-norm statistical convergence rate of de-sparsified Graphical Lasso could be easily derived from the $\ell_\infty$-norm rate. In this way, we compare CRDA and \TheName\ through the $\ell_2$-norm statistical convergence rate of their inverse covariance matrix estimators.
Thus, we can derive the $\ell_2$-norm convergence rate as:
\begin{equation}
\begin{aligned}
\|\widehat T-\Theta^{*}\|_2 = \mathcal{O}\left(\sqrt{\frac{d  \log d}{N}}\right).
\end{aligned}
\label{eq:conv}
\end{equation}
On the other side,~\cite{rothman2008sparse} demonstrated that the $\ell_2$-norm convergence rate for the Graphical Lasso estimator $\widehat\Theta$ is
$$\|\widehat\Theta-\Theta^*\|_2=\mathcal{O}\left(\sqrt{\frac{(d + s) \log d}{N}}\right),$$
where $s$ has been defined in \textbf{Corollary 3}. 
%i.e.,
%
%$$d=.$$
%
We conclude \emph{the convergence rate of $\widehat T$ is faster than $\widehat\Theta$}. In this way, we consider \TheName\ would outperform CRDA as it adopts a better inverse covariance matrix estimator.

\section{Experiments}
In this section, we first introduce the design of the experiments to evaluate the superiority of the proposed \TheName{} framework. Then, we present the experimental results, including the performance comparison between the \TheName{} framework, existing LDA baselines, and other predictive models, followed by a comparison between inverse covariance matrix to support our theoretical analysis of \TheName.

\subsection{Experiment Setups}
 In this study, to evaluate \TheName{}, we use predictive analytics of disease based on Electronic Health Records (EHR) data.
\begin{itemize}
    
\item  {\bf Predictive Analytics of Diseases -}   Given $N$ training samples (i.e., the EHR data of each patient) along with corresponding labels i.e., $(x_1,l_1),(x_2,l_2)\dots$ $(x_N,l_N)$ where $l_i\in\{-1,+1\}$ refers to whether the patient $i$ is diagnosed with the target disease or not (i.e., positive sample or negative sample), the predictive analytics task is to determine whether a new patient would develop into the target disease via classification of the vector $x$ to $+1$ (diagnosed as the positive result) or $-1$ (diagnosed as the negative result).

\item{\bf Performance Metrics - }    To demonstrate the effectiveness of predictive analytics of diseases, we compared all these methods with other competitors using  \emph{\bf Accuracy} and \emph{\bf F1-score}.  Specifically, Accuracy characterizes the proportion of patients who are accurately classified by the algorithms. F1-Score measures both correctness and completeness of the prediction. Of-course, we also include other metrics such as sensitivity and specificity to evaluate the performance of predictive analytics through addressing the medical concerns.

\item \textbf{Data for Evaluation - }   In the experiments, we use the de-identified EHR data of 200,000 students from ten U.S. universities~\cite{turner_college_2015}.  Among all diseases recorded, we choose mental health disorders, including \emph{anxiety disorders, mood disorders, depression disorders, and other related disorders}, as one targeted disease for early detection~\cite{kendler2003life}.  We represent each patient using his/her diagnosis-frequency vector based on the clustered codeset ($d=295$).

\end{itemize}
Note that, to prepare the training and testing sets, we use the complete EHR data of the patients who haven't been diagnosed with any of mental health disorders (negative samples). For patients having been diagnosed with any mental health disorders (positive samples), we collect their EHR data from the first visit to the last visit that was 90 days before the diagnosis of mental health disorders. Thus, we can simulate the early detection of diseases with 90 days in advance.

\subsection{Design of Experiment}
To understand the performance impact of \TheName{} beyond classic LDA, we first propose four LDA baseline approaches to compare against \TheName{}, then, three discriminative learning models are used for the comparison:
%\begin{itemize}
\begin{itemize}
%\item
    \item LDA Derivatives: \emph{LDA}, \emph{Shrinkage}, \emph{DIAG} and \emph{Ye-LDA}  -- The first three algorithms are all based on the common implementation of generalized Fishier's discriminant analysis listed in Eq. ~\ref{eq:gLDA}. Specifically, \emph{LDA} uses the sample covariance estimation, and inverts the covariance matrix using pseudo-inverse~\cite{ye2004optimization} when the matrix inverse is not available; \emph{Shrinkage} is based on \emph{LDA}, using a sparse estimation of sample covariance as follows,
    \begin{equation}
        \Sigma_\beta=\beta*\bar\Sigma+(1-\beta) * \mathrm{diag}(\bar\Sigma)\ \text{and}\ \Theta_\beta=\Sigma_\beta^{-1},
    \end{equation}
    where $\mathrm{diag}(\bar\Sigma)$ refers to the diagonal matrix of the sample estimation $\bar\Sigma$. \emph{DIAG} is a special \emph{Shrinkage approach} with $\beta=0.0$. Ye-LDA is derived from~\cite{Ye2004,ye2004optimization}. In our research, we focus on studying the improvement  of LDA classification caused by (inverse) covariance matrix regularization, thus we don't compare our method to linear-coefficient-regularized LDA classifiers~\cite{clemmensen2011sparse,shao2011sparse,buhlmann2011statistics} or heuristic LDA derivation~\cite{conf/eccv/BelhumeurHK96}.

%\item
\item Downstream Classifiers: \emph{Support Vector Machine (SVM),  Logistic Regression (Logit. Reg.) and AdaBoost} --  Inspired by the previous studies~\cite{personalized2015,huang2014toward} in EHR data mining, we use a linear binary SVM classifier with fine-tuned parameters as well as the Logistic Regression classifier. Further, we compare our algorithm to AdaBoost, where AdaBoost-10 refers to the AdaBoost classifier using 10 Logistic Regression instances and AdaBoost-50 leverages 50 instances.
%\end{itemize}
%
\end{itemize}
With the seven baseline algorithms, we perform experiments with training sets of varying sizes and cross-validation. To train the classifiers, we randomly selected 50, 100, 150, 200, and 250  positive patients, and randomly selected the same number of negative patients. Then, we test the classifiers, using a testing set with 1000 randomly selected positive patients and the same number of negative patients. Note that there is no over-lap between training set and the paired testing set. All algorithms used in our work are implemented with JSAT~\footnote{https://github.com/EdwardRaff/JSAT} and glasso in R~\footnote{https://cran.r-project.org/package=glasso}.

%For each setting, we run the eight algorithms and repeat 30 times.

\subsection{Overall Comparison}

We include the comparison results of \TheName{} evaluation in Tables~\ref{tab:50pc},~\ref{tab:100pc},~\ref{tab:150pc},~\ref{tab:200pc}, and~\ref{tab:250pc} for models learned from $50\sim250\times 2$ labeled samples respectively. All experiments are done with cross validation using random sampling without replacement and repeated 10 times.
Specifically, we compare the performance using various experimental settings, such as the varying
parameters for model training and number of days in advance for early detection(e.g., 30 days, 60 days and 90 days). We carry out the experiments with varying \emph{Days in Advance} settings, so as to evaluate the performance of algorithms for predictive analytics. As was addressed, we actually need to use the past EHR data (before the diagnoses of mental disorders) as the features for prediction. More specific, for positive samples in both training and testing datasets, we backtracked their EHR data from their prediction dates. For every positive sample, the prediction date is set as 30, 60 and 90 days before the medicare visit that the patient received his/her first diagnoses of ``anxiety disorders, mood disorders, depression disorders, and other related disorders''. In this way, we carry out the experiments in three categories according to the varying \emph{days in advance}.

\subsubsection{Comparisons on Accuracy and F1-Score}
As can be seen from the results in Tables~\ref{tab:50pc},~\ref{tab:100pc},~\ref{tab:150pc},~\ref{tab:200pc}, and~\ref{tab:250pc}, \TheName\ clearly outperforms the baseline algorithms in terms of overall accuracy and F1-score.
Specifically, \TheName{} achieves 18.6\%--21.3\% increase in accuracy and 22.9\%--32\% increase in F1-score over LDA;
\TheName{} achieves 17.9\% increase in accuracy and 31.5\%--40.6\% increase in F1-score over DIAG.
Compared to Shrinkage and CRDA, the accuracy and F1-score of \TheName{} in most parameter settings are 0.3\%--18.9\% higher and 0.14\%--71.8\% higher, respectively.
%Compared to those robust classifiers such as SVM, Logistic Regression, and AdaBoost, \TheName{} still clearly outperforms these baseline algorithms.
%
Compared to SVM, Logistic Regression, and AdaBoost, \TheName{} can achieve 2.3\%--19.4\% higher accuracy and 7.5\%--43.5\% higher F1-score.  In this case, we can conclude that the classic LDA model cannot perform as well as many other predictive models such as SVM and AdaBoost. However, \TheName{} significantly outperforms these methods in all settings.
Thus, we can conclude that \TheName\ overall outperforms the baseline algorithms in all experimental settings. Note that, though  \TheName{} outperforms CRDA marginally, \TheName{} enjoys a more tight upper bound of error rate.

%\subsubsection{Overall Comparison }
%Tables~\ref{tab:50pc} and~\ref{tab:250pc} report the performance of \TheName{} and baselines under various settings.
% on $1000\time 2$ testing samples, while all other results are attached in the Appendix B.

%Additional comparisons considering different training and testing set sizes as well as different days in advance (30, 60, and 90 days) are provided in Appendix.

%

\begin{table*}
\caption{Performance Comparison with Training Set:50$\times$ 2, Testing Set: 2000$\times$2}
\label{tab:50pc}
\centering\small
\begin{tabular}{*{5}{l}}
\toprule
 & Accuracy & F1-Score & Sensitivity & Specificity\\
\hline\multicolumn{5}{c}{  Days in Advance: 30}\\\hline
AdaBoost ($\times 10$)&0.637 $\pm$ 0.028&0.571 $\pm$ 0.057&0.491 $\pm$ 0.085&0.783 $\pm$ 0.053\\
AdaBoost ($\times 50$)&0.640 $\pm$ 0.024&0.570 $\pm$ 0.061&0.487 $\pm$ 0.093&0.792 $\pm$ 0.053\\
CRDA ($\lambda=1.0$)&0.662 $\pm$ 0.017&0.692 $\pm$ 0.028&0.762 $\pm$ 0.069&0.563 $\pm$ 0.058\\
CRDA ($\lambda=10.0$)&0.670 $\pm$ 0.017&0.713 $\pm$ 0.010&0.819 $\pm$ 0.023&0.520 $\pm$ 0.047\\
CRDA ($\lambda=100.0$)&0.664 $\pm$ 0.020&0.713 $\pm$ 0.008&0.834 $\pm$ 0.033&0.494 $\pm$ 0.068\\
LDA&0.555 $\pm$ 0.026&0.565 $\pm$ 0.033&0.579 $\pm$ 0.048&0.531 $\pm$ 0.040\\
Logistic Regression &0.615 $\pm$ 0.055&0.469 $\pm$ 0.206&0.395 $\pm$ 0.200&0.835 $\pm$ 0.094\\
\TheName{} ($\lambda=1.0$)&0.658 $\pm$ 0.019&0.677 $\pm$ 0.034&0.723 $\pm$ 0.073&0.592 $\pm$ 0.050\\
\TheName{} ($\lambda=10.0$)&\textbf{0.672 $\pm$ 0.015} &0.713 $\pm$ 0.010&0.813 $\pm$ 0.025&0.532 $\pm$ 0.042\\
\TheName{} ($\lambda=100.0$)&0.668 $\pm$ 0.018&\textbf{0.714 $\pm$ 0.008}&0.830 $\pm$ 0.026&0.506 $\pm$ 0.056\\
SVM&0.611 $\pm$ 0.026&0.619 $\pm$ 0.034&0.632 $\pm$ 0.050&0.590 $\pm$ 0.029\\
DIAG&0.568 $\pm$ 0.014&0.515 $\pm$ 0.026&0.460 $\pm$ 0.042&0.676 $\pm$ 0.046\\
Shrinkage ($\beta=0.25$)&0.574 $\pm$ 0.014&0.538 $\pm$ 0.025&0.499 $\pm$ 0.041&0.649 $\pm$ 0.045\\
Shrinkage ($\beta=0.5$)&0.560 $\pm$ 0.033&0.438 $\pm$ 0.220&0.413 $\pm$ 0.210&0.708 $\pm$ 0.152\\
Shrinkage ($\beta=0.75$)&0.560 $\pm$ 0.025&0.480 $\pm$ 0.163&0.448 $\pm$ 0.158&0.672 $\pm$ 0.118\\
\hline\multicolumn{5}{c}{  Days in Advance: 60}\\\hline
AdaBoost ($\times 10$)&0.646 $\pm$ 0.021&0.596 $\pm$ 0.054&0.531 $\pm$ 0.095&0.762 $\pm$ 0.057\\
AdaBoost ($\times 50$)&0.639 $\pm$ 0.027&0.569 $\pm$ 0.083&0.491 $\pm$ 0.111&0.788 $\pm$ 0.060\\
CRDA ($\lambda=1.0$)&0.654 $\pm$ 0.016&0.690 $\pm$ 0.016&0.774 $\pm$ 0.067&0.535 $\pm$ 0.088\\
CRDA ($\lambda=10.0$)&0.653 $\pm$ 0.019&0.706 $\pm$ 0.010&0.833 $\pm$ 0.053&0.474 $\pm$ 0.083\\
CRDA ($\lambda=100.0$)&0.643 $\pm$ 0.024&0.701 $\pm$ 0.028&0.844 $\pm$ 0.098&0.443 $\pm$ 0.124\\
LDA&0.556 $\pm$ 0.028&0.550 $\pm$ 0.042&0.547 $\pm$ 0.072&0.565 $\pm$ 0.065\\
Logistic Regression &0.631 $\pm$ 0.031&0.535 $\pm$ 0.108&0.447 $\pm$ 0.132&0.814 $\pm$ 0.073\\
\TheName{} ($\lambda=1.0$)&0.655 $\pm$ 0.012&0.675 $\pm$ 0.023&0.723 $\pm$ 0.070&0.587 $\pm$ 0.074\\
\TheName{} ($\lambda=10.0$)&\textbf{0.661 $\pm$ 0.016}&\textbf{0.708 $\pm$ 0.009}&0.823 $\pm$ 0.051&0.499 $\pm$ 0.077\\
\TheName{} ($\lambda=100.0$)&0.649 $\pm$ 0.021&0.705 $\pm$ 0.020&0.844 $\pm$ 0.082&0.454 $\pm$ 0.110\\
SVM&0.627 $\pm$ 0.019&0.625 $\pm$ 0.027&0.625 $\pm$ 0.053&0.629 $\pm$ 0.056\\
DIAG&0.565 $\pm$ 0.011&0.514 $\pm$ 0.046&0.468 $\pm$ 0.076&0.662 $\pm$ 0.072\\
Shrinkage ($\beta=0.25$)&0.568 $\pm$ 0.012&0.530 $\pm$ 0.040&0.492 $\pm$ 0.069&0.644 $\pm$ 0.063\\
Shrinkage ($\beta=0.5$)&0.567 $\pm$ 0.013&0.528 $\pm$ 0.038&0.489 $\pm$ 0.067&0.646 $\pm$ 0.059\\
Shrinkage ($\beta=0.75$)&0.561 $\pm$ 0.025&0.477 $\pm$ 0.164&0.444 $\pm$ 0.163&0.677 $\pm$ 0.120\\
\hline\multicolumn{5}{c}{  Days in Advance: 90}\\\hline
AdaBoost ($\times 10$)&0.627 $\pm$ 0.034&0.572 $\pm$ 0.063&0.507 $\pm$ 0.091&0.747 $\pm$ 0.054\\
AdaBoost ($\times 50$)&0.632 $\pm$ 0.035&0.575 $\pm$ 0.054&0.504 $\pm$ 0.077&0.759 $\pm$ 0.058\\
CRDA ($\lambda=1.0$)&0.641 $\pm$ 0.018&0.663 $\pm$ 0.041&0.716 $\pm$ 0.106&0.566 $\pm$ 0.091\\
CRDA ($\lambda=10.0$)&0.651 $\pm$ 0.018&0.693 $\pm$ 0.034&0.797 $\pm$ 0.093&0.505 $\pm$ 0.096\\
CRDA ($\lambda=100.0$)&0.634 $\pm$ 0.040&0.675 $\pm$ 0.101&0.808 $\pm$ 0.188&0.459 $\pm$ 0.173\\
LDA&0.546 $\pm$ 0.025&0.532 $\pm$ 0.038&0.518 $\pm$ 0.058&0.574 $\pm$ 0.046\\
Logistic Regression &0.597 $\pm$ 0.058&0.423 $\pm$ 0.217&0.351 $\pm$ 0.207&0.843 $\pm$ 0.096\\
\TheName{} ($\lambda=1.0$)&0.642 $\pm$ 0.022&0.663 $\pm$ 0.035&0.710 $\pm$ 0.078&0.574 $\pm$ 0.060\\
\TheName{} ($\lambda=10.0$)&\textbf{0.658 $\pm$ 0.016}&\textbf{0.696 $\pm$ 0.022}&0.787 $\pm$ 0.073&0.528 $\pm$ 0.084\\
\TheName{} ($\lambda=100.0$)&0.641 $\pm$ 0.031&0.683 $\pm$ 0.081&0.808 $\pm$ 0.164&0.475 $\pm$ 0.148\\
SVM&0.597 $\pm$ 0.034&0.600 $\pm$ 0.036&0.606 $\pm$ 0.047&0.587 $\pm$ 0.046\\
DIAG&0.568 $\pm$ 0.023&0.514 $\pm$ 0.048&0.464 $\pm$ 0.074&0.672 $\pm$ 0.066\\
Shrinkage ($\beta=0.25$)&0.569 $\pm$ 0.020&0.530 $\pm$ 0.041&0.490 $\pm$ 0.065&0.648 $\pm$ 0.054\\
Shrinkage ($\beta=0.5$)&0.565 $\pm$ 0.021&0.519 $\pm$ 0.041&0.473 $\pm$ 0.059&0.657 $\pm$ 0.044\\
Shrinkage ($\beta=0.75$)&0.559 $\pm$ 0.019&0.511 $\pm$ 0.040&0.465 $\pm$ 0.061&0.653 $\pm$ 0.050\\
\bottomrule
\end{tabular}
\end{table*}

\begin{table*}
\caption{Performance Comparison with Training Set:100$\times$ 2, Testing Set: 2000$\times$2}
\label{tab:100pc}
\centering\small
\begin{tabular}{*{5}{l}}
\toprule
 & Accuracy & F1-Score & Sensitivity & Specificity\\
\hline\multicolumn{5}{c}{  Days in Advance: 30}\\\hline
AdaBoost ($\times 10$)&0.632 $\pm$ 0.029&0.541 $\pm$ 0.095&0.452 $\pm$ 0.117&0.812 $\pm$ 0.065\\
AdaBoost ($\times 50$)&0.631 $\pm$ 0.032&0.538 $\pm$ 0.099&0.447 $\pm$ 0.120&0.814 $\pm$ 0.062\\
CRDA ($\lambda=1.0$)&0.674 $\pm$ 0.012&0.708 $\pm$ 0.019&0.792 $\pm$ 0.043&0.556 $\pm$ 0.029\\
CRDA ($\lambda=10.0$)&0.675 $\pm$ 0.006&0.722 $\pm$ 0.008&0.844 $\pm$ 0.022&0.507 $\pm$ 0.017\\
CRDA ($\lambda=100.0$)&0.664 $\pm$ 0.010&0.718 $\pm$ 0.004&0.858 $\pm$ 0.031&0.469 $\pm$ 0.048\\
LDA&0.594 $\pm$ 0.016&0.592 $\pm$ 0.019&0.591 $\pm$ 0.027&0.597 $\pm$ 0.018\\
Logistic Regression &0.593 $\pm$ 0.054&0.394 $\pm$ 0.200&0.305 $\pm$ 0.180&0.881 $\pm$ 0.075\\
\TheName{} ($\lambda=1.0$)&0.674 $\pm$ 0.018&0.700 $\pm$ 0.025&0.765 $\pm$ 0.050&0.582 $\pm$ 0.026\\
\TheName{} ($\lambda=10.0$)&\textbf{0.681} $\pm$ 0.006&\textbf{0.724 $\pm$ 0.006}&0.838 $\pm$ 0.018&0.524 $\pm$ 0.020\\
\TheName{} ($\lambda=100.0$)&0.668 $\pm$ 0.009&0.720 $\pm$ 0.006&0.854 $\pm$ 0.028&0.481 $\pm$ 0.041\\
SVM&0.636 $\pm$ 0.016&0.642 $\pm$ 0.024&0.655 $\pm$ 0.044&0.618 $\pm$ 0.025\\
DIAG&0.594 $\pm$ 0.019&0.562 $\pm$ 0.034&0.524 $\pm$ 0.050&0.663 $\pm$ 0.033\\
Shrinkage ($\beta=0.25$)&0.600 $\pm$ 0.020&0.582 $\pm$ 0.031&0.559 $\pm$ 0.045&0.641 $\pm$ 0.022\\
Shrinkage ($\beta=0.5$)&0.581 $\pm$ 0.044&0.467 $\pm$ 0.235&0.449 $\pm$ 0.228&0.714 $\pm$ 0.144\\
Shrinkage ($\beta=0.75$)&0.599 $\pm$ 0.014&0.582 $\pm$ 0.020&0.559 $\pm$ 0.029&0.639 $\pm$ 0.022\\
\hline\multicolumn{5}{c}{  Days in Advance: 60}\\\hline
AdaBoost ($\times 10$)&0.633 $\pm$ 0.024&0.537 $\pm$ 0.076&0.439 $\pm$ 0.110&0.827 $\pm$ 0.067\\
AdaBoost ($\times 50$)&0.623 $\pm$ 0.024&0.507 $\pm$ 0.065&0.396 $\pm$ 0.089&0.850 $\pm$ 0.052\\
CRDA ($\lambda=1.0$)&0.676 $\pm$ 0.016&0.711 $\pm$ 0.015&0.797 $\pm$ 0.041&0.555 $\pm$ 0.052\\
CRDA ($\lambda=10.0$)&0.672 $\pm$ 0.019&0.719 $\pm$ 0.015&0.837 $\pm$ 0.025&0.508 $\pm$ 0.039\\
CRDA ($\lambda=100.0$)&0.668 $\pm$ 0.017&0.716 $\pm$ 0.013&0.838 $\pm$ 0.038&0.498 $\pm$ 0.054\\
LDA&0.603 $\pm$ 0.024&0.599 $\pm$ 0.026&0.595 $\pm$ 0.033&0.610 $\pm$ 0.032\\
Logistic Regression &0.613 $\pm$ 0.042&0.462 $\pm$ 0.164&0.362 $\pm$ 0.147&0.863 $\pm$ 0.069\\
\TheName{} ($\lambda=1.0$)&\textbf{0.679 $\pm$ 0.011}&0.707 $\pm$ 0.014&0.776 $\pm$ 0.041&0.582 $\pm$ 0.043\\
\TheName{} ($\lambda=10.0$)&0.676 $\pm$ 0.016&\textbf{0.720 $\pm$ 0.012}&0.834 $\pm$ 0.026&0.518 $\pm$ 0.039\\
\TheName{} ($\lambda=100.0$)&0.671 $\pm$ 0.017&0.718 $\pm$ 0.012&0.838 $\pm$ 0.029&0.504 $\pm$ 0.045\\
SVM&0.644 $\pm$ 0.016&0.645 $\pm$ 0.020&0.650 $\pm$ 0.038&0.637 $\pm$ 0.037\\
DIAG&0.596 $\pm$ 0.015&0.562 $\pm$ 0.033&0.522 $\pm$ 0.058&0.670 $\pm$ 0.054\\
Shrinkage ($\beta=0.25$)&0.600 $\pm$ 0.016&0.580 $\pm$ 0.024&0.554 $\pm$ 0.040&0.645 $\pm$ 0.038\\
Shrinkage ($\beta=0.5$)&0.596 $\pm$ 0.035&0.532 $\pm$ 0.178&0.513 $\pm$ 0.174&0.680 $\pm$ 0.113\\
Shrinkage ($\beta=0.75$)&0.596 $\pm$ 0.039&0.532 $\pm$ 0.179&0.513 $\pm$ 0.175&0.678 $\pm$ 0.115\\
\hline\multicolumn{5}{c}{  Days in Advance: 90}\\\hline
AdaBoost ($\times 10$)&0.626 $\pm$ 0.022&0.519 $\pm$ 0.061&0.412 $\pm$ 0.093&0.840 $\pm$ 0.058\\
AdaBoost ($\times 50$)&0.631 $\pm$ 0.017&0.523 $\pm$ 0.056&0.413 $\pm$ 0.087&0.849 $\pm$ 0.053\\
CRDA ($\lambda=1.0$)&0.674 $\pm$ 0.013&0.709 $\pm$ 0.020&0.796 $\pm$ 0.052&0.552 $\pm$ 0.047\\
CRDA ($\lambda=10.0$)&0.674 $\pm$ 0.010&0.721 $\pm$ 0.006&0.845 $\pm$ 0.021&0.502 $\pm$ 0.034\\
CRDA ($\lambda=100.0$)&0.666 $\pm$ 0.015&0.719 $\pm$ 0.006&0.856 $\pm$ 0.025&0.477 $\pm$ 0.052\\
LDA&0.605 $\pm$ 0.017&0.607 $\pm$ 0.026&0.612 $\pm$ 0.045&0.598 $\pm$ 0.028\\
Logistic Regression &0.611 $\pm$ 0.036&0.453 $\pm$ 0.130&0.345 $\pm$ 0.136&0.876 $\pm$ 0.067\\
\TheName{} ($\lambda=1.0$)&0.675 $\pm$ 0.013&0.700 $\pm$ 0.026&0.764 $\pm$ 0.061&0.587 $\pm$ 0.045\\
\TheName{} ($\lambda=10.0$)&\textbf{0.682 $\pm$ 0.007}&\textbf{0.725 $\pm$ 0.007}&0.840 $\pm$ 0.025&0.523 $\pm$ 0.030\\
\TheName{} ($\lambda=100.0$)&0.669 $\pm$ 0.013&0.721 $\pm$ 0.006&0.853 $\pm$ 0.023&0.486 $\pm$ 0.046\\
SVM&0.632 $\pm$ 0.017&0.638 $\pm$ 0.023&0.649 $\pm$ 0.039&0.616 $\pm$ 0.026\\
DIAG&0.597 $\pm$ 0.015&0.574 $\pm$ 0.039&0.549 $\pm$ 0.072&0.644 $\pm$ 0.063\\
Shrinkage ($\beta=0.25$)&0.593 $\pm$ 0.034&0.531 $\pm$ 0.179&0.517 $\pm$ 0.182&0.668 $\pm$ 0.120\\
Shrinkage ($\beta=0.5$)&0.602 $\pm$ 0.015&0.589 $\pm$ 0.028&0.575 $\pm$ 0.053&0.628 $\pm$ 0.043\\
Shrinkage ($\beta=0.75$)&0.599 $\pm$ 0.015&0.586 $\pm$ 0.025&0.570 $\pm$ 0.045&0.629 $\pm$ 0.037\\
\bottomrule
\end{tabular}
\end{table*}

\begin{table*}
\caption{Performance Comparison with Training Set:150$\times$ 2, Testing Set: 2000$\times$2}
\label{tab:150pc}
\centering\small
\begin{tabular}{*{5}{l}}
\toprule
 & Accuracy & F1-Score & Sensitivity & Specificity\\
\hline\multicolumn{5}{c}{  Days in Advance: 30}\\\hline
AdaBoost ($\times 10$)&0.615 $\pm$ 0.010&0.484 $\pm$ 0.033&0.363 $\pm$ 0.039&0.867 $\pm$ 0.024\\
AdaBoost ($\times 50$)&0.615 $\pm$ 0.007&0.482 $\pm$ 0.025&0.359 $\pm$ 0.032&0.871 $\pm$ 0.023\\
CRDA ($\lambda=1.0$)&\textbf{0.682 $\pm$ 0.008}&0.723 $\pm$ 0.008&0.829 $\pm$ 0.021&0.534 $\pm$ 0.019\\
CRDA ($\lambda=10.0$)&0.671 $\pm$ 0.013&0.721 $\pm$ 0.008&0.851 $\pm$ 0.016&0.490 $\pm$ 0.035\\
CRDA ($\lambda=100.0$)&0.662 $\pm$ 0.014&0.718 $\pm$ 0.007&0.861 $\pm$ 0.020&0.464 $\pm$ 0.044\\
LDA&0.613 $\pm$ 0.012&0.611 $\pm$ 0.018&0.610 $\pm$ 0.038&0.615 $\pm$ 0.037\\
Logistic Regression &0.581 $\pm$ 0.045&0.352 $\pm$ 0.189&0.255 $\pm$ 0.142&0.908 $\pm$ 0.053\\
\TheName{} ($\lambda=1.0$)&0.681 $\pm$ 0.009&0.712 $\pm$ 0.012&0.790 $\pm$ 0.028&0.572 $\pm$ 0.020\\
\TheName{} ($\lambda=10.0$)&0.681 $\pm$ 0.007&\textbf{0.727 $\pm$ 0.006}&0.849 $\pm$ 0.013&0.512 $\pm$ 0.019\\
\TheName{} ($\lambda=100.0$)&0.667 $\pm$ 0.013&0.720 $\pm$ 0.007&0.857 $\pm$ 0.020&0.478 $\pm$ 0.041\\
SVM&0.650 $\pm$ 0.012&0.660 $\pm$ 0.014&0.680 $\pm$ 0.024&0.620 $\pm$ 0.023\\
DIAG&0.619 $\pm$ 0.014&0.610 $\pm$ 0.031&0.600 $\pm$ 0.056&0.637 $\pm$ 0.037\\
Shrinkage ($\beta=0.25$)&0.599 $\pm$ 0.051&0.500 $\pm$ 0.251&0.503 $\pm$ 0.256&0.696 $\pm$ 0.156\\
Shrinkage ($\beta=0.5$)&0.611 $\pm$ 0.039&0.562 $\pm$ 0.189&0.566 $\pm$ 0.195&0.656 $\pm$ 0.121\\
Shrinkage ($\beta=0.75$)&0.615 $\pm$ 0.009&0.611 $\pm$ 0.024&0.608 $\pm$ 0.051&0.623 $\pm$ 0.045\\
\hline\multicolumn{5}{c}{  Days in Advance: 60}\\\hline
AdaBoost ($\times 10$)&0.625 $\pm$ 0.039&0.512 $\pm$ 0.131&0.424 $\pm$ 0.156&0.826 $\pm$ 0.081\\
AdaBoost ($\times 50$)&0.637 $\pm$ 0.024&0.554 $\pm$ 0.072&0.466 $\pm$ 0.113&0.809 $\pm$ 0.068\\
CRDA ($\lambda=1.0$)&0.677 $\pm$ 0.017&0.717 $\pm$ 0.015&0.818 $\pm$ 0.028&0.536 $\pm$ 0.032\\
CRDA ($\lambda=10.0$)&0.671 $\pm$ 0.012&0.721 $\pm$ 0.008&0.848 $\pm$ 0.022&0.494 $\pm$ 0.038\\
CRDA ($\lambda=100.0$)&0.662 $\pm$ 0.014&0.718 $\pm$ 0.006&0.861 $\pm$ 0.031&0.463 $\pm$ 0.055\\
LDA&0.623 $\pm$ 0.014&0.621 $\pm$ 0.023&0.619 $\pm$ 0.040&0.627 $\pm$ 0.023\\
Logistic Regression &0.600 $\pm$ 0.054&0.412 $\pm$ 0.217&0.331 $\pm$ 0.195&0.869 $\pm$ 0.090\\
\TheName{} ($\lambda=1.0$)&\textbf{0.681 $\pm$ 0.016}&0.711 $\pm$ 0.016&0.787 $\pm$ 0.033&0.574 $\pm$ 0.036\\
\TheName{} ($\lambda=10.0$)&0.678 $\pm$ 0.011&\textbf{0.724 $\pm$ 0.009}&0.843 $\pm$ 0.017&0.513 $\pm$ 0.023\\
\TheName{} ($\lambda=100.0$)&0.667 $\pm$ 0.014&0.720 $\pm$ 0.007&0.856 $\pm$ 0.028&0.477 $\pm$ 0.050\\
SVM&0.649 $\pm$ 0.017&0.654 $\pm$ 0.025&0.665 $\pm$ 0.042&0.633 $\pm$ 0.024\\
DIAG&0.615 $\pm$ 0.018&0.597 $\pm$ 0.032&0.574 $\pm$ 0.054&0.656 $\pm$ 0.045\\
Shrinkage ($\beta=0.25$)&0.618 $\pm$ 0.018&0.605 $\pm$ 0.031&0.587 $\pm$ 0.051&0.649 $\pm$ 0.039\\
Shrinkage ($\beta=0.5$)&0.608 $\pm$ 0.039&0.548 $\pm$ 0.184&0.533 $\pm$ 0.181&0.683 $\pm$ 0.110\\
Shrinkage ($\beta=0.75$)&0.618 $\pm$ 0.015&0.602 $\pm$ 0.027&0.581 $\pm$ 0.045&0.655 $\pm$ 0.033\\
\hline\multicolumn{5}{c}{  Days in Advance: 90}\\\hline
AdaBoost ($\times 10$)&0.630 $\pm$ 0.023&0.531 $\pm$ 0.075&0.436 $\pm$ 0.123&0.824 $\pm$ 0.082\\
AdaBoost ($\times 50$)&0.630 $\pm$ 0.023&0.534 $\pm$ 0.078&0.441 $\pm$ 0.126&0.820 $\pm$ 0.083\\
CRDA ($\lambda=1.0$)&0.674 $\pm$ 0.012&0.708 $\pm$ 0.017&0.794 $\pm$ 0.045&0.553 $\pm$ 0.039\\
CRDA ($\lambda=10.0$)&0.671 $\pm$ 0.011&0.720 $\pm$ 0.007&0.845 $\pm$ 0.021&0.498 $\pm$ 0.035\\
CRDA ($\lambda=100.0$)&0.663 $\pm$ 0.013&0.718 $\pm$ 0.004&0.857 $\pm$ 0.025&0.470 $\pm$ 0.050\\
LDA&0.611 $\pm$ 0.020&0.610 $\pm$ 0.025&0.608 $\pm$ 0.039&0.614 $\pm$ 0.024\\
Logistic Regression &0.614 $\pm$ 0.045&0.463 $\pm$ 0.174&0.374 $\pm$ 0.180&0.853 $\pm$ 0.098\\
\TheName{} ($\lambda=1.0$)&0.672 $\pm$ 0.018&0.693 $\pm$ 0.030&0.745 $\pm$ 0.065&0.600 $\pm$ 0.042\\
\TheName{} ($\lambda=10.0$)&\textbf{0.678 $\pm$ 0.010}&\textbf{0.722 $\pm$ 0.009}&0.836 $\pm$ 0.026&0.521 $\pm$ 0.033\\
\TheName{} ($\lambda=100.0$)&0.668 $\pm$ 0.010&0.720 $\pm$ 0.005&0.851 $\pm$ 0.022&0.485 $\pm$ 0.039\\
SVM&0.639 $\pm$ 0.015&0.645 $\pm$ 0.020&0.657 $\pm$ 0.035&0.622 $\pm$ 0.026\\
DIAG&0.610 $\pm$ 0.012&0.602 $\pm$ 0.022&0.590 $\pm$ 0.042&0.631 $\pm$ 0.031\\
Shrinkage ($\beta=0.25$)&0.613 $\pm$ 0.011&0.608 $\pm$ 0.019&0.601 $\pm$ 0.036&0.626 $\pm$ 0.027\\
Shrinkage ($\beta=0.5$)&0.602 $\pm$ 0.036&0.547 $\pm$ 0.183&0.540 $\pm$ 0.183&0.665 $\pm$ 0.114\\
Shrinkage ($\beta=0.75$)&0.601 $\pm$ 0.036&0.545 $\pm$ 0.183&0.536 $\pm$ 0.182&0.665 $\pm$ 0.113\\
\bottomrule
\end{tabular}
\end{table*}

\begin{table*}
\caption{Performance Comparison with Training Set:200$\times$ 2, Testing Set: 2000$\times$2}
\label{tab:200pc}
\centering\small
\begin{tabular}{*{5}{l}}
\toprule
 & Accuracy & F1-Score & Sensitivity & Specificity\\
\hline\multicolumn{5}{c}{  Days in Advance: 30}\\\hline
AdaBoost ($\times 10$)&0.618 $\pm$ 0.026&0.485 $\pm$ 0.082&0.373 $\pm$ 0.115&0.863 $\pm$ 0.064\\
AdaBoost ($\times 50$)&0.618 $\pm$ 0.022&0.491 $\pm$ 0.064&0.377 $\pm$ 0.092&0.859 $\pm$ 0.052\\
CRDA ($\lambda=1.0$)&\textbf{0.688 $\pm$ 0.006}&0.725 $\pm$ 0.007&0.824 $\pm$ 0.017&0.553 $\pm$ 0.016\\
CRDA ($\lambda=10.0$)&0.680 $\pm$ 0.005&0.725 $\pm$ 0.005&0.847 $\pm$ 0.013&0.513 $\pm$ 0.013\\
CRDA ($\lambda=100.0$)&0.669 $\pm$ 0.011&0.721 $\pm$ 0.003&0.855 $\pm$ 0.026&0.483 $\pm$ 0.047\\
LDA&0.637 $\pm$ 0.006&0.644 $\pm$ 0.010&0.655 $\pm$ 0.021&0.620 $\pm$ 0.020\\
Logistic Regression &0.598 $\pm$ 0.046&0.411 $\pm$ 0.175&0.313 $\pm$ 0.159&0.883 $\pm$ 0.070\\
\TheName{} ($\lambda=1.0$)&0.686 $\pm$ 0.007&0.717 $\pm$ 0.007&0.794 $\pm$ 0.017&0.578 $\pm$ 0.019\\
\TheName{} ($\lambda=10.0$)&0.684 $\pm$ 0.006&\textbf{0.729 $\pm$ 0.005}&0.850 $\pm$ 0.007&0.519 $\pm$ 0.010\\
\TheName{} ($\lambda=100.0$)&0.673 $\pm$ 0.009&0.723 $\pm$ 0.004&0.852 $\pm$ 0.024&0.494 $\pm$ 0.038\\
SVM&0.660 $\pm$ 0.012&0.671 $\pm$ 0.012&0.693 $\pm$ 0.014&0.626 $\pm$ 0.015\\
DIAG&0.623 $\pm$ 0.013&0.603 $\pm$ 0.024&0.575 $\pm$ 0.041&0.671 $\pm$ 0.029\\
Shrinkage ($\beta=0.25$)&0.628 $\pm$ 0.013&0.621 $\pm$ 0.023&0.610 $\pm$ 0.039&0.646 $\pm$ 0.024\\
Shrinkage ($\beta=0.5$)&0.619 $\pm$ 0.042&0.565 $\pm$ 0.190&0.560 $\pm$ 0.190&0.678 $\pm$ 0.110\\
Shrinkage ($\beta=0.75$)&0.633 $\pm$ 0.012&0.629 $\pm$ 0.019&0.624 $\pm$ 0.034&0.642 $\pm$ 0.022\\
\hline\multicolumn{5}{c}{  Days in Advance: 60}\\\hline
AdaBoost ($\times 10$)&0.605 $\pm$ 0.023&0.445 $\pm$ 0.085&0.325 $\pm$ 0.074&0.885 $\pm$ 0.033\\
AdaBoost ($\times 50$)&0.616 $\pm$ 0.010&0.479 $\pm$ 0.038&0.356 $\pm$ 0.048&0.876 $\pm$ 0.032\\
CRDA ($\lambda=1.0$)&\textbf{0.684 $\pm$ 0.006}&0.721 $\pm$ 0.006&0.818 $\pm$ 0.019&0.549 $\pm$ 0.023\\
CRDA ($\lambda=10.0$)&0.674 $\pm$ 0.008&0.722 $\pm$ 0.006&0.844 $\pm$ 0.019&0.505 $\pm$ 0.026\\
CRDA ($\lambda=100.0$)&0.673 $\pm$ 0.010&0.721 $\pm$ 0.006&0.845 $\pm$ 0.021&0.502 $\pm$ 0.035\\
LDA&0.626 $\pm$ 0.009&0.622 $\pm$ 0.013&0.616 $\pm$ 0.028&0.635 $\pm$ 0.031\\
Logistic Regression &0.589 $\pm$ 0.038&0.380 $\pm$ 0.151&0.270 $\pm$ 0.113&0.908 $\pm$ 0.038\\
\TheName{} ($\lambda=1.0$)&0.684 $\pm$ 0.010&0.710 $\pm$ 0.012&0.773 $\pm$ 0.027&0.595 $\pm$ 0.023\\
\TheName{} ($\lambda=10.0$)&0.682 $\pm$ 0.006&\textbf{0.726 $\pm$ 0.007}&0.844 $\pm$ 0.017&0.520 $\pm$ 0.014\\
\TheName{} ($\lambda=100.0$)&0.675 $\pm$ 0.008&0.722 $\pm$ 0.006&0.843 $\pm$ 0.022&0.508 $\pm$ 0.031\\
SVM&0.651 $\pm$ 0.006&0.659 $\pm$ 0.010&0.675 $\pm$ 0.026&0.626 $\pm$ 0.028\\
DIAG&0.627 $\pm$ 0.012&0.615 $\pm$ 0.023&0.597 $\pm$ 0.045&0.657 $\pm$ 0.039\\
Shrinkage ($\beta=0.25$)&0.618 $\pm$ 0.041&0.562 $\pm$ 0.188&0.553 $\pm$ 0.187&0.683 $\pm$ 0.111\\
Shrinkage ($\beta=0.5$)&0.620 $\pm$ 0.040&0.565 $\pm$ 0.189&0.557 $\pm$ 0.187&0.683 $\pm$ 0.110\\
Shrinkage ($\beta=0.75$)&0.616 $\pm$ 0.039&0.557 $\pm$ 0.186&0.544 $\pm$ 0.183&0.688 $\pm$ 0.109\\
\hline\multicolumn{5}{c}{  Days in Advance: 90}\\\hline
AdaBoost ($\times 10$)&0.626 $\pm$ 0.033&0.507 $\pm$ 0.107&0.411 $\pm$ 0.153&0.840 $\pm$ 0.088\\
AdaBoost ($\times 50$)&0.632 $\pm$ 0.028&0.533 $\pm$ 0.092&0.441 $\pm$ 0.135&0.823 $\pm$ 0.080\\
CRDA ($\lambda=1.0$)&0.682 $\pm$ 0.008&0.722 $\pm$ 0.008&0.825 $\pm$ 0.017&0.540 $\pm$ 0.020\\
CRDA ($\lambda=10.0$)&0.664 $\pm$ 0.012&0.718 $\pm$ 0.006&0.856 $\pm$ 0.025&0.472 $\pm$ 0.044\\
CRDA ($\lambda=100.0$)&0.656 $\pm$ 0.016&0.715 $\pm$ 0.005&0.865 $\pm$ 0.029&0.447 $\pm$ 0.058\\
LDA&0.631 $\pm$ 0.014&0.630 $\pm$ 0.018&0.631 $\pm$ 0.034&0.630 $\pm$ 0.032\\
Logistic Regression &0.605 $\pm$ 0.060&0.424 $\pm$ 0.232&0.353 $\pm$ 0.222&0.857 $\pm$ 0.107\\
\TheName{} ($\lambda=1.0$)&\textbf{0.684 $\pm$ 0.010}&0.714 $\pm$ 0.014&0.789 $\pm$ 0.031&0.579 $\pm$ 0.020\\
\TheName{} ($\lambda=10.0$)&0.676 $\pm$ 0.008&\textbf{0.724 $\pm$ 0.004}&0.852 $\pm$ 0.019&0.500 $\pm$ 0.030\\
\TheName{} ($\lambda=100.0$)&0.658 $\pm$ 0.015&0.716 $\pm$ 0.005&0.863 $\pm$ 0.029&0.452 $\pm$ 0.057\\
SVM&0.657 $\pm$ 0.009&0.669 $\pm$ 0.015&0.693 $\pm$ 0.031&0.621 $\pm$ 0.024\\
DIAG&0.625 $\pm$ 0.013&0.614 $\pm$ 0.029&0.601 $\pm$ 0.055&0.648 $\pm$ 0.045\\
Shrinkage ($\beta=0.25$)&0.627 $\pm$ 0.014&0.617 $\pm$ 0.030&0.604 $\pm$ 0.056&0.651 $\pm$ 0.043\\
Shrinkage ($\beta=0.5$)&0.626 $\pm$ 0.013&0.616 $\pm$ 0.027&0.603 $\pm$ 0.051&0.650 $\pm$ 0.042\\
Shrinkage ($\beta=0.75$)&0.626 $\pm$ 0.014&0.617 $\pm$ 0.023&0.604 $\pm$ 0.045&0.649 $\pm$ 0.043\\
\bottomrule
\end{tabular}
\end{table*}

\begin{table*}
\caption{Performance Comparison with Training Set:250$\times$ 2, Testing Set: 2000$\times$2}
\label{tab:250pc}
\centering\small
\begin{tabular}{*{5}{l}}
\toprule
 & Accuracy & F1-Score & Sensitivity & Specificity\\
\hline\multicolumn{5}{c}{  Days in Advance: 30}\\\hline
AdaBoost ($\times 10$)&0.620 $\pm$ 0.037&0.484 $\pm$ 0.110&0.380 $\pm$ 0.147&0.860 $\pm$ 0.076\\
AdaBoost ($\times 50$)&0.625 $\pm$ 0.033&0.499 $\pm$ 0.097&0.394 $\pm$ 0.138&0.856 $\pm$ 0.074\\
CRDA ($\lambda=1.0$)&0.689 $\pm$ 0.010&0.726 $\pm$ 0.009&0.824 $\pm$ 0.021&0.553 $\pm$ 0.025\\
CRDA ($\lambda=10.0$)&0.677 $\pm$ 0.012&0.722 $\pm$ 0.009&0.840 $\pm$ 0.020&0.513 $\pm$ 0.029\\
CRDA ($\lambda=100.0$)&0.666 $\pm$ 0.014&0.719 $\pm$ 0.007&0.853 $\pm$ 0.027&0.479 $\pm$ 0.050\\
LDA&0.644 $\pm$ 0.009&0.645 $\pm$ 0.012&0.648 $\pm$ 0.023&0.640 $\pm$ 0.020\\
Logistic Regression &0.605 $\pm$ 0.057&0.424 $\pm$ 0.204&0.339 $\pm$ 0.200&0.870 $\pm$ 0.089\\
\TheName{} ($\lambda=1.0$)&\textbf{0.690 $\pm$ 0.007}&0.719 $\pm$ 0.007&0.791 $\pm$ 0.022&0.589 $\pm$ 0.027\\
\TheName{} ($\lambda=10.0$)&0.684 $\pm$ 0.009&\textbf{0.726 $\pm$ 0.008}&0.837 $\pm$ 0.015&0.531 $\pm$ 0.012\\
\TheName{} ($\lambda=100.0$)&0.671 $\pm$ 0.012&0.721 $\pm$ 0.008&0.848 $\pm$ 0.026&0.494 $\pm$ 0.039\\
SVM&0.663 $\pm$ 0.013&0.673 $\pm$ 0.015&0.694 $\pm$ 0.024&0.632 $\pm$ 0.023\\
DIAG&0.633 $\pm$ 0.011&0.619 $\pm$ 0.028&0.599 $\pm$ 0.055&0.668 $\pm$ 0.046\\
Shrinkage ($\beta=0.25$)&0.625 $\pm$ 0.044&0.569 $\pm$ 0.192&0.562 $\pm$ 0.193&0.689 $\pm$ 0.108\\
Shrinkage ($\beta=0.5$)&0.626 $\pm$ 0.044&0.569 $\pm$ 0.192&0.561 $\pm$ 0.192&0.691 $\pm$ 0.106\\
Shrinkage ($\beta=0.75$)&0.639 $\pm$ 0.011&0.633 $\pm$ 0.022&0.624 $\pm$ 0.039&0.653 $\pm$ 0.025\\
\hline\multicolumn{5}{c}{  Days in Advance: 60}\\\hline
AdaBoost ($\times 10$)&0.635 $\pm$ 0.026&0.539 $\pm$ 0.087&0.449 $\pm$ 0.141&0.820 $\pm$ 0.091\\
AdaBoost ($\times 50$)&0.634 $\pm$ 0.027&0.536 $\pm$ 0.089&0.445 $\pm$ 0.144&0.823 $\pm$ 0.091\\
CRDA ($\lambda=1.0$)&0.692 $\pm$ 0.006&0.729 $\pm$ 0.006&0.827 $\pm$ 0.014&0.557 $\pm$ 0.015\\
CRDA ($\lambda=10.0$)&0.682 $\pm$ 0.008&0.730 $\pm$ 0.004&0.860 $\pm$ 0.019&0.504 $\pm$ 0.031\\
CRDA ($\lambda=100.0$)&0.674 $\pm$ 0.014&0.726 $\pm$ 0.005&0.864 $\pm$ 0.025&0.483 $\pm$ 0.051\\
LDA&0.642 $\pm$ 0.011&0.643 $\pm$ 0.015&0.645 $\pm$ 0.025&0.638 $\pm$ 0.017\\
Logistic Regression &0.623 $\pm$ 0.048&0.489 $\pm$ 0.184&0.411 $\pm$ 0.195&0.835 $\pm$ 0.105\\
\TheName{} ($\lambda=1.0$)&\textbf{0.691 $\pm$ 0.008}&0.717 $\pm$ 0.009&0.781 $\pm$ 0.019&0.601 $\pm$ 0.017\\
\TheName{} ($\lambda=10.0$)&0.689 $\pm$ 0.004&\textbf{0.733 $\pm$ 0.004}&0.854 $\pm$ 0.013&0.524 $\pm$ 0.017\\
\TheName{} ($\lambda=100.0$)&0.676 $\pm$ 0.013&0.727 $\pm$ 0.005&0.863 $\pm$ 0.024&0.488 $\pm$ 0.048\\
SVM&0.662 $\pm$ 0.008&0.668 $\pm$ 0.012&0.681 $\pm$ 0.023&0.642 $\pm$ 0.017\\
DIAG&0.634 $\pm$ 0.012&0.613 $\pm$ 0.026&0.582 $\pm$ 0.053&0.687 $\pm$ 0.049\\
Shrinkage ($\beta=0.25$)&0.627 $\pm$ 0.044&0.565 $\pm$ 0.189&0.545 $\pm$ 0.185&0.709 $\pm$ 0.101\\
Shrinkage ($\beta=0.5$)&0.642 $\pm$ 0.010&0.634 $\pm$ 0.015&0.620 $\pm$ 0.028&0.663 $\pm$ 0.028\\
Shrinkage ($\beta=0.75$)&0.641 $\pm$ 0.010&0.636 $\pm$ 0.012&0.627 $\pm$ 0.021&0.655 $\pm$ 0.022\\
\hline\multicolumn{5}{c}{  Days in Advance: 90}\\\hline
AdaBoost ($\times 10$)&0.633 $\pm$ 0.027&0.536 $\pm$ 0.089&0.447 $\pm$ 0.140&0.818 $\pm$ 0.086\\
AdaBoost ($\times 50$)&0.631 $\pm$ 0.026&0.535 $\pm$ 0.087&0.445 $\pm$ 0.137&0.818 $\pm$ 0.085\\
CRDA ($\lambda=1.0$)&0.686 $\pm$ 0.006&0.721 $\pm$ 0.009&0.813 $\pm$ 0.029&0.558 $\pm$ 0.026\\
CRDA ($\lambda=10.0$)&0.675 $\pm$ 0.007&0.720 $\pm$ 0.006&0.838 $\pm$ 0.021&0.512 $\pm$ 0.028\\
CRDA ($\lambda=100.0$)&0.671 $\pm$ 0.009&0.719 $\pm$ 0.004&0.844 $\pm$ 0.028&0.497 $\pm$ 0.043\\
LDA&0.648 $\pm$ 0.009&0.648 $\pm$ 0.018&0.651 $\pm$ 0.037&0.644 $\pm$ 0.025\\
Logistic Regression &0.628 $\pm$ 0.028&0.520 $\pm$ 0.095&0.427 $\pm$ 0.146&0.828 $\pm$ 0.090\\
\TheName{} ($\lambda=1.0$)&\textbf{0.687 $\pm$ 0.009}&0.713 $\pm$ 0.014&0.778 $\pm$ 0.033&0.597 $\pm$ 0.022\\
\TheName{} ($\lambda=10.0$)&0.683 $\pm$ 0.006&\textbf{0.725 $\pm$ 0.008}&0.839 $\pm$ 0.021&0.527 $\pm$ 0.018\\
\TheName{} ($\lambda=100.0$)&0.673 $\pm$ 0.008&0.720 $\pm$ 0.005&0.841 $\pm$ 0.024&0.505 $\pm$ 0.037\\
SVM&0.666 $\pm$ 0.009&0.672 $\pm$ 0.014&0.687 $\pm$ 0.030&0.644 $\pm$ 0.023\\
DIAG&0.635 $\pm$ 0.015&0.621 $\pm$ 0.030&0.601 $\pm$ 0.053&0.668 $\pm$ 0.032\\
Shrinkage ($\beta=0.25$)&0.638 $\pm$ 0.012&0.631 $\pm$ 0.027&0.621 $\pm$ 0.051&0.656 $\pm$ 0.032\\
Shrinkage ($\beta=0.5$)&0.642 $\pm$ 0.011&0.635 $\pm$ 0.026&0.626 $\pm$ 0.050&0.657 $\pm$ 0.032\\
Shrinkage ($\beta=0.75$)&0.641 $\pm$ 0.010&0.635 $\pm$ 0.024&0.628 $\pm$ 0.046&0.655 $\pm$ 0.030\\
\bottomrule
\end{tabular}
\end{table*}

%\subsubsection{Comparisons under Medicare Settings}
%In this section, 

\subsubsection{Trade-off between Sensitivity and Specificity}
We also intend to compare \TheName\ with baseline methods with respect to the needs of medicines. Specifically, in addition to accuracy and F1-score, we focus on two more metrics~\cite{altman1994diagnostic}:
\begin{itemize}
    \item \emph{Sensitivity - } In medical diagnosis, the sensitivity measures the ability of the prediction algorithms to \emph{correctly identify the patients with the disease (true positive rate)}. More specific, we estimate sensitivity as
    \begin{equation}
        \mathrm{Sensitivity}=\frac{\text{\# Patients with the diseases} \cap \text{Patients predicted as positive} }{\text{\#Patients with the diseases}}.
    \end{equation}
    \item \emph{Specificity - } In contrast, the specificity metric characterizes the ability of the algorithms to \emph{correctly identify ones without the disease (true negative rate)}. More specific, we estimate specificity as
    \begin{equation}
        \mathrm{Specificity}=\frac{\text{\# Patients without the diseases} \cap \text{Patients predicted as negative} }{\text{\#Patients without the diseases}}.
    \end{equation}
\end{itemize}
%
%
%\TheName{} clearly outperforms other algorithms in terms of overall accuracy, F1-score and sensitivity in all settings. 
Please see also Tables~\ref{tab:50pc},~\ref{tab:100pc},~\ref{tab:150pc},~\ref{tab:200pc}, and~\ref{tab:250pc}. In terms of specificity, the baseline algorithms outperform \TheName{}, in
the most of cases. However, in terms of sensitivity and specificity trade-off, \TheName{} on average gains
19.5\% higher sensitivity while sacrificing 8.2\% specificity, when compared to typical LDA. On opposite side of the trade-off, when compared to CRDA (based on graphical lasso), \TheName{} on average gains
2.3\% higher specificity while sacrificing 1.4\% sensitivity. %Please refer to the Appendix for more comparison.
%Thus, 

\subsubsection{Discussion on the Performance Comparison}
We consider testing accuracy and F1-score as two primary metrics for the evaluation, as these two metrics well characterize the performance of classifiers. Thus, we conclude that \TheName{} overall outperforms the baseline algorithms, including both LDA, SVM, Logistic Regression, and other classifiers, in all experimental settings. In terms of the trade-off between sensitivity and specificity, we argue that \TheName\ still outperforms the original LDA classifier and CRDA classifiers, considering the requests of predictive analytics of diseases and the early diagnosis. While the original LDA classifier well-balances the sensitivity and specificity, both CRDA and \TheName\ would incorporate slightly higher sensitivity, compared to the original LDA, while having lower specificity. In this way, CRDA and \TheName\ could discover more patients potentially with the diseases, but also slightly raise the frequency of false alarms. We believe, compared to the marginal increase of false alarms, the improvement of sensitivity should be appreciated in medical contexts. Compared \TheName\ to CRDA, the de-sparsified Graphical Lasso here helps \TheName\ achieve higher overall accuracy and F1-score with a more balanced pair of sensitivity and specificity.

\begin{figure*}
%\vspace{-5mm}
\centering
     \subfloat[Error Reduction of All Estimators Beyond Sample Estimation]{\includegraphics[width=0.7\textwidth]{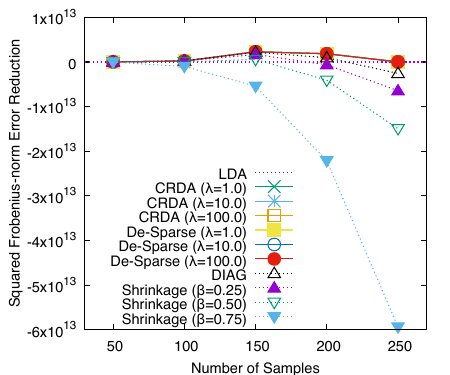}}\\
     \subfloat[Zoom-in on \TheName\ vs. CRDA]{\includegraphics[width=0.7\textwidth]{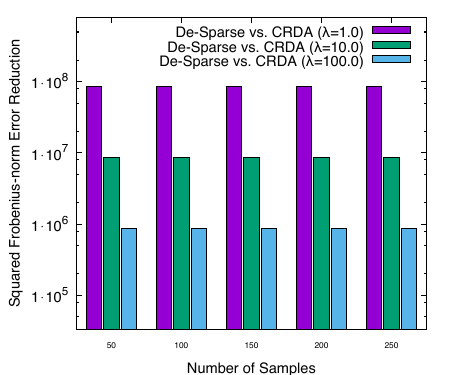}}
    % \vspace{-3mm}
     \caption{Comparison on $\ell_2$-norm Estimator Error Reduction of Inverse Covariance Matrices Estimation (the higher the better)}
\label{fig:error-est}
%\vspace{-5mm}
\end{figure*}

\subsection{Empirical Convergence of Parameter Estimation}

We hypothesize \TheName{} improves LDA because that the de-sparsified Graphical Lasso used in \TheName{} approaches to the inverse of population covariance matrix has a tighter error bound than the inverse of sample covariance matrix used in simple LDA models, when the training sample size is limited.
In order to verify our hypothesis, we compare the inverse covariance matrix estimators used in \TheName, CRDA, and other LDA baselines, using the EHR data. Specifically, we (1) learned a ``ground truth'' covariance matrix $\Sigma_\mathrm{GT}$ (and its inverse $\Theta_\mathrm{GT}=\Sigma_\mathrm{GT}^{-1}$) using diagnosis-frequency vectors of $10,000$ patients (w/o the target disease, balanced) randomly retrieved from all patients of 22 U.S. university healthcare systems, (2) randomly selected another $50$ to $250$ samples (w/o the target disease, balanced) to train LDA, \TheName{}, CRDA and Shrinkage, (3) estimated the error between the inverse covariance matrix (denoted as $\Theta$, $\Theta=\widehat\Theta$ for CRDA, $\Theta=\widehat T$ for \TheName, and $\Theta=\Theta_\beta$ for Shrinkage LDA) learned in each classifier \emph{versus} the inverse of ``ground truth'' covariance matrix $\Sigma_\mathrm{GT}$, all in $\ell_2$-norm,
%:
%
%$$e{rr}(\widehat\Sigma^{-1})=||\Sigma_l^{-1}-\widehat\Sigma^{-1}||_2^2,$$
%ß
and (4) further estimated the \emph{error reduction} of $\Theta$ from the inverse of sample covariance estimation (i.e., $\bar\Theta=\bar\Sigma^{-1}$) as
\begin{equation}
    R(\Theta) = \|\Theta-\bar\Theta\|_2^2-\|\Theta_\mathrm{GT}-\Theta\|_2^2,
\end{equation}
where $\Theta=\widehat\Theta$ for CRDA, $\Theta=\widehat T$ for \TheName, and $\Theta=\Theta_\beta$ for Shrinkage LDA. We repeated above (1)--(4) steps for 100 times, and illustrated the average \emph{error reduction} $R(\Theta)$ in Fig.~\ref{fig:error-est}(a), with varying parameters and settings.
%
% Note that, positive error reduction means the corresponding estimator outperforms the sample estimation with \emph{lower estimation error}, while negative error reduction means the estimator performs even poorer than the sample estimation.

Fig.~\ref{fig:error-est}(a) demonstrates that, estimators used in \TheName\ ($\widehat T$) and CRDA ($\widehat\Theta$) outperform the sample estimation in all settings, while DIAG and Shrinkage  estimators (i.e., $\Theta_{\beta}$ and $\beta=0.0$, $0.25,$ $0.5,$ and $0.75$) may cause even \emph{higher estimation error} (with negative error reduction) when the number of samples increases.
Fig.~\ref{fig:error-est}(b) illustrates the trend of error reduction with CRDA and \TheName. Though the difference between these two algorithms is not visible in such scale, we can observe that these two algorithms achieve the maximal error reduction when number of samples is $150$ in our experiments, while the error reduction is low when the number of samples is relatively small (50) or large (250). Because, when the sample size is small, both sample-based estimation ($\bar\Theta$) and the regularized estimation ($\widehat T$ and $\widehat\Theta$) work poorly, though $\widehat T$ and $\widehat\Theta$ still outperform $\bar\Theta$. With the increasing sample size, the advantage of CRDA and \TheName{} becomes more and more significant. However, when sample size is large, both sample-based estimation and the regularized estimation converge well, thus the error reduction becomes marginal.

\section{Conclusion}

%There are many existing approaches to represent EHR data including the use of diagnosis-frequencies~\cite{sun2012supervised,7091853,personalized2015}, pairwise diagnosis transition~\cite{jensen2001mining}, and graph representations of diagnosis sequences~\cite{liu_temporal_2015}. Among these approaches, the diagnosis-frequency is a common way to represent EHR data.

In this paper, we study the long existing problem of covariance-regularized discriminant analysis for classification under high-dimensional low sample sizes (HDLSS) settings. More specific, we take care of the applications to the predictive analytics of diseases using Electronic Health Records (EHRs) data and common diagnosis-frequency data representation.
To understand the performance of LDA, we extend the existing theory~\cite{lachenbruch1968estimation,zollanvari2013random} and propose a novel analytical model characterizing the error rate of LDA classification under the uncertainty of parameter estimation.  Based on the analytical model, we propose \TheName{} -- a novel LDA classifier using de-sparsified Graphical Lasso. Our analysis shows that the proposed algorithm could outperform the existing Covariance-regularized discriminant analysis (CRDA) based on common Graphical Lasso.
The experimental results on real-world Electronic Health Record (EHR) datasets show \TheName{} outperforms all baseline algorithms. We interpret the comparison of results and demonstrate the advantage of proposed methods in medicare settings. Further, the empirical studies on estimator comparison validate our analysis.

%Note that, there exists other sparse inverse covariance matrix estimators~\cite{cai2016estimating,wang2016precision} with even faster convergence rate under various sparsity conditions. Due to the limit of contents, we don't further introduce them. These estimators might be able to improve the performance of covariance-regularized discriminant analysis, and the philosophy behind coincides our theory.

%\small{
\bibliographystyle{unsrt}
\bibliography{submission}
%}
%\include{proof}
%\include{appendix}

\end{document}